\documentclass[10pt,twocolumn,letterpaper]{article}

\usepackage{iccv}
\usepackage{times}
\usepackage{epsfig}
\usepackage{graphicx}
\usepackage{amsmath}
\usepackage{amssymb}

\usepackage{tabularx}
\usepackage{booktabs}
\usepackage{rotating}
\usepackage{multirow}
\usepackage{threeparttable}

\usepackage{subcaption}
\usepackage{caption}
\DeclareMathOperator*{\argmax}{\arg\max}
\DeclareMathOperator*{\argmin}{\arg\min}
\usepackage{xcolor}
\newcommand{\gray}[1]{\textcolor{gray}{#1}}
\usepackage{wrapfig}

\iccvfinalcopy 

\ificcvfinal\fi

\begin{document}
	
	\title{Unsupervised Video Anomaly Detection\\ via Normalizing Flows with Implicit Latent Features}
	
	\author{MyeongAh Cho$^{1}$ \hspace{1.0em} Taeoh Kim$^{2}$ \hspace{1.0em} Woo Jin Kim$^1$ \hspace{1.0em} Suhwan Cho$^1$ \hspace{1.0em} Sangyoun Lee$^{1}$\thanks{\textit{Published on Pattern Recognition, Elsevier}}\\
		\\
		$^1$Yonsei University, Korea\\
		$^2$NAVER CLOVA Video, Korea\\
	}
	
	\maketitle
	\ificcvfinal\thispagestyle{empty}\fi
	
	\begin{abstract}
		In contemporary society, surveillance anomaly detection, i.e., spotting anomalous events such as crimes or accidents in surveillance videos, is a critical task. As anomalies occur rarely, most training data consists of unlabeled videos without anomalous events, which makes the task challenging. Most existing methods use an autoencoder (AE) to learn to reconstruct normal videos; they then detect anomalies based on their failure to reconstruct the appearance of abnormal scenes. However, because anomalies are distinguished by appearance as well as motion, many previous approaches have explicitly separated appearance and motion information—for example, using a pre-trained optical flow model. This explicit separation restricts reciprocal representation capabilities between two types of information. In contrast, we propose an implicit two-path AE (ITAE), a structure in which two encoders implicitly model appearance and motion features, along with a single decoder that combines them to learn normal video patterns. For the complex distribution of normal scenes, we suggest normal density estimation of ITAE features through normalizing flow (NF)-based generative models to learn the tractable likelihoods and identify anomalies using out-of-distribution detection. NF models intensify ITAE performance by learning normality through implicitly learned features. Finally, we demonstrate the effectiveness of ITAE and its feature distribution modeling on six benchmarks, including databases that contain various anomalies in real-world scenarios.
	\end{abstract}
	
	\section{Introduction}
	\begin{figure}[!t]
		\centering
		\subfloat[Abnormal frame]{\includegraphics[width=0.3\columnwidth]{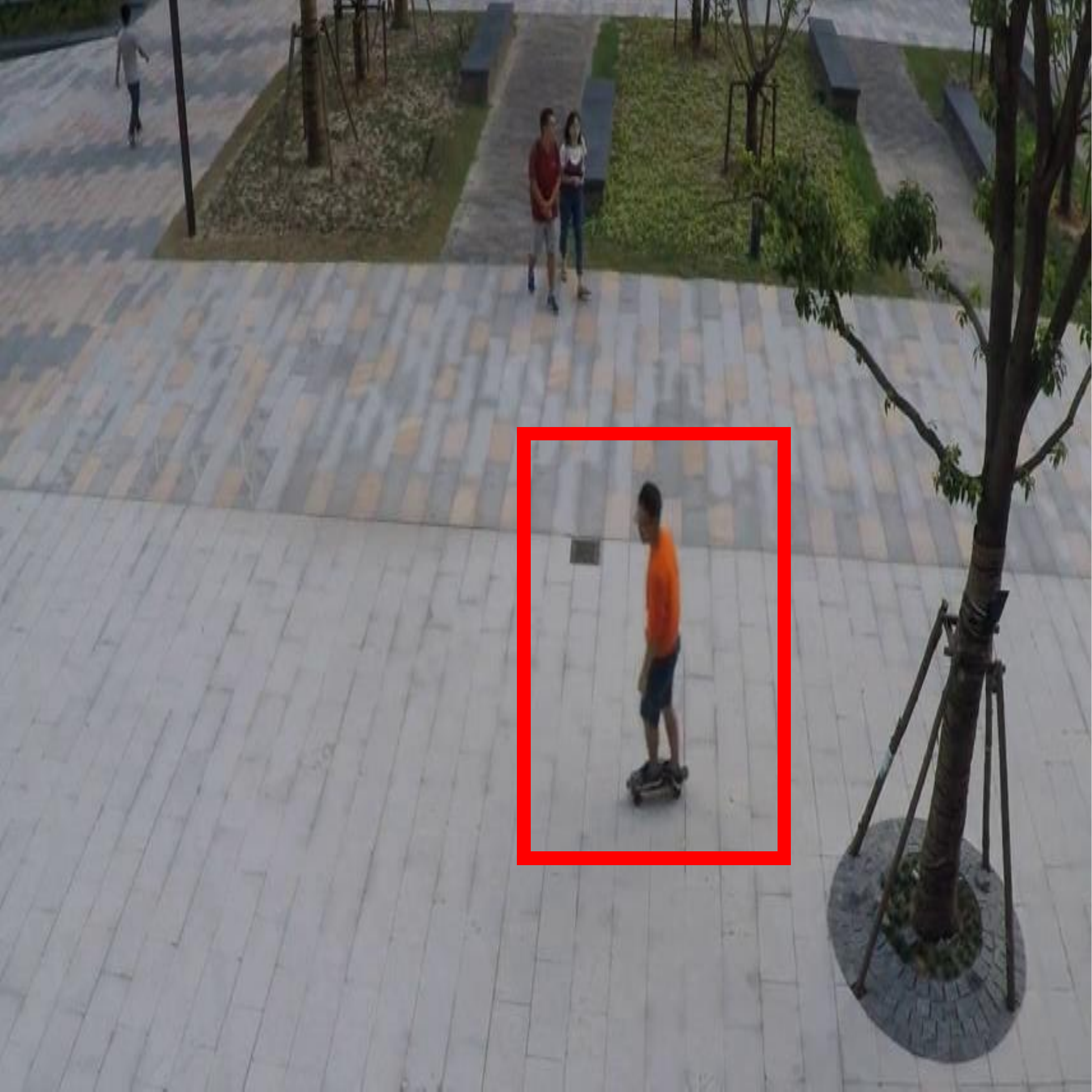} \label{f0_a}}
		\hspace{0.2em}
		\subfloat[AE]{\includegraphics[width=0.3\columnwidth]{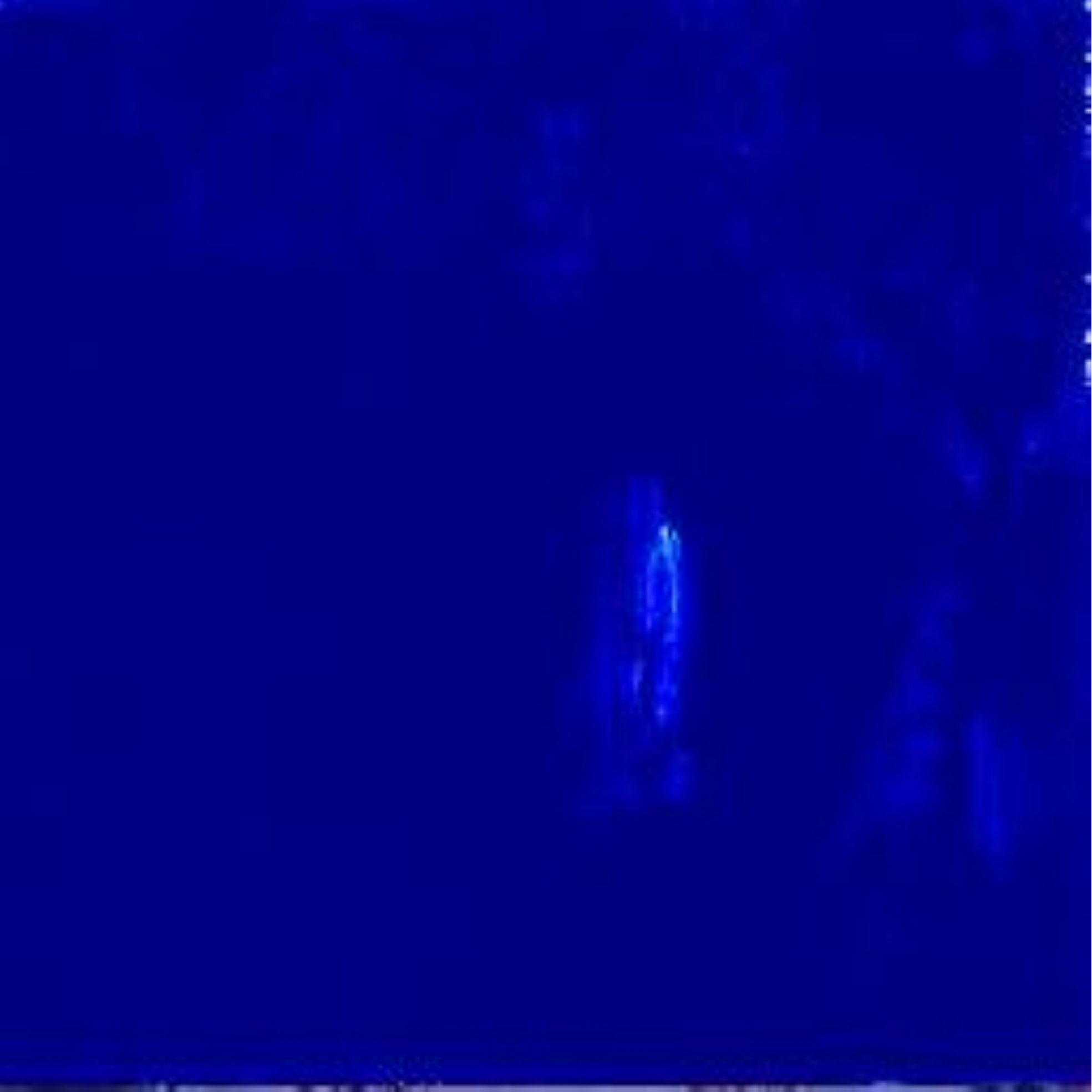}\label{f0_b}}
		\hspace{0.2em}
		\subfloat[ITAE]{\includegraphics[width=0.3\columnwidth]{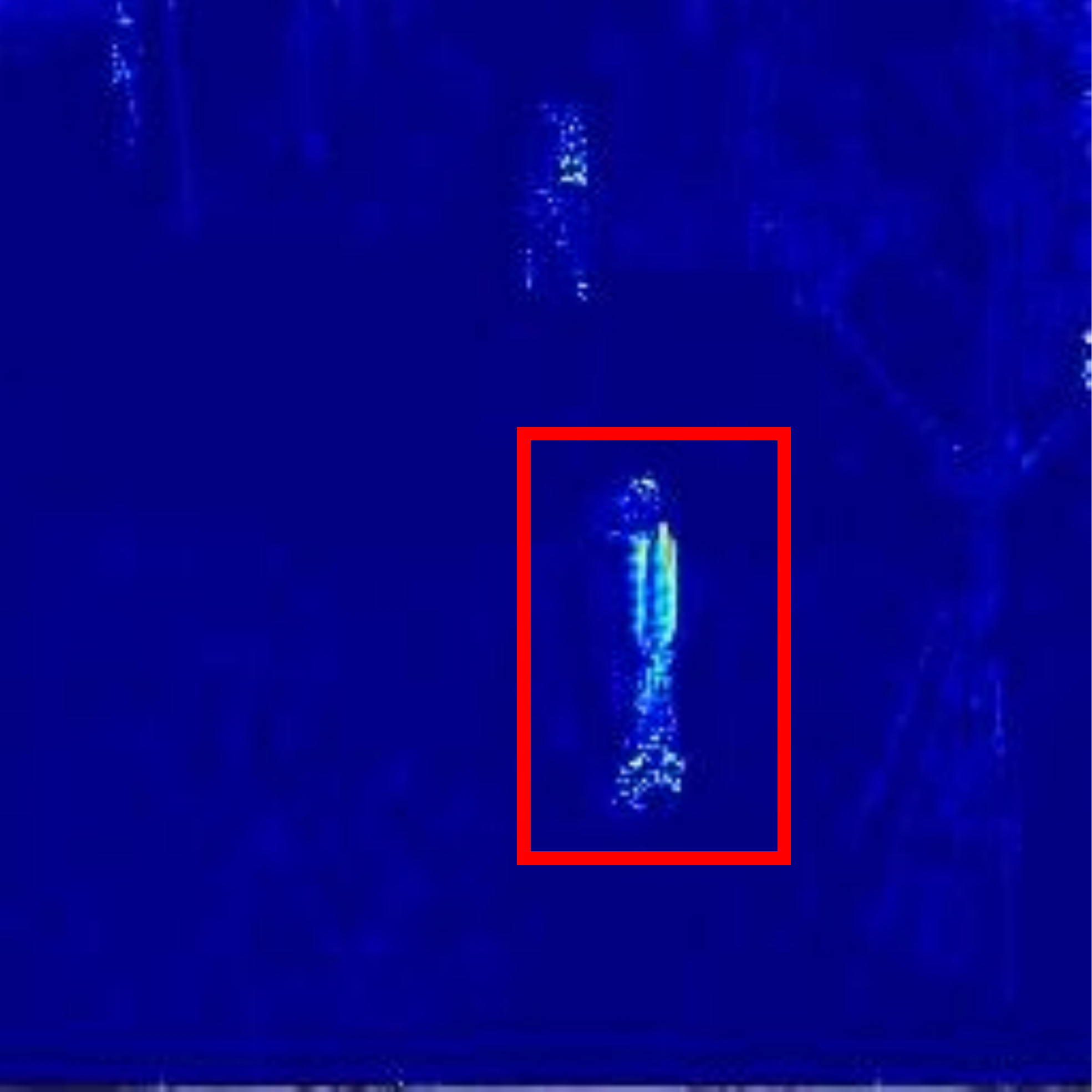}\label{f0_c}}
		\\
		\subfloat[Distribution of normal and abnormal frames
		]{\includegraphics[width=0.98\columnwidth]{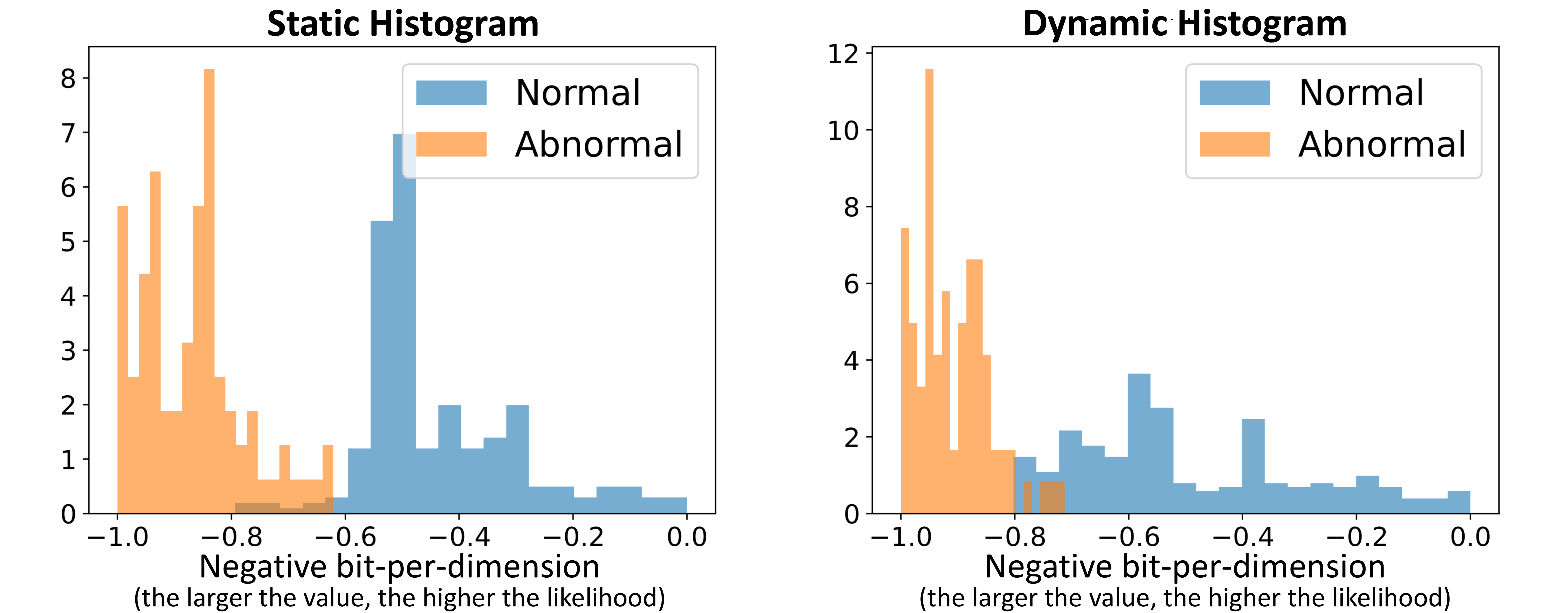}\label{f0_d}}
		\caption{Reconstruction error of (a) abnormal input from (b) AE and the proposed (c) ITAE, and (d) log-likelihood histogram of video clip. For a scene where a person rides a skateboard, where the motion factor is heavily abnormal, AE with only a static encoder reconstructs accurately, which fails to detect anomalies. In contrast, ITAE with static and dynamic encoder produces large errors, and each normality distribution histogram of the latent feature enables distinguishing anomalies.} 
		\label{f1}
	\end{figure}
	
	Anomaly detection, also called outlier detection, seeks to identify unusual, unseen, or undefined abnormal data among normal data. Anomaly detection has several practical applications in various fields such as surveillance anomaly detection, defect detection in factories, X-ray security systems, and medical diagnostics. At present, with CCTVs present in most places, video anomaly detection such as detection of accidents and crimes from amongst the petabytes of surveillance videos has become critical. Furthermore, human monitoring for unpredictable anomalous events tends to be time-consuming, laborious, and error-prone, and must be replaced with an automated intelligent system. 
	
	However, there are several challenges. First, real-world anomalous events such as robberies and car accidents occur very infrequently compared with normal events, resulting in a class imbalance problem between normal and abnormal data. Therefore, the training sets of most surveillance databases only contain normal videos, while anomalous events only exist in the test set. This makes it challenging to train models in a general supervised manner that uses manually labeled data. Second, because anomalies are unbounded, it is impossible to define and collect all existing abnormal events, and the task of labeling is extremely laborious. Therefore, detecting unseen and undefined anomalous events requires the system to learn normality through abundant and easily obtained normal videos. 
	
	Since the advent of deep learning, studies on surveillance anomaly detection tasks with a large number of normal training videos have grown substantially. Frame reconstruction or prediction-based methods are used predominantly in unsupervised learning approaches~\cite{tang2020integrating, liu2018future}. Autoencoder (AE) structured networks that learn reconstruction (or prediction) tasks with only normal scenes cannot reconstruct properly when abnormal scenes are input during testing, entailing large reconstruction errors between the input and output for anomaly detection. This approach enables training without labeled data and has achieved notable improvement in performance.
	
	In a surveillance system, anomalous events can be distinguished from normal events based on appearance, motion, or both. For example, the presence of non-pedestrian objects such as cars traversing on the sidewalk has a different appearance from a normal scene; fighting or chasing people illustrate differences in motion; and people throwing abnormal objects exhibit differences in both. It is essential to extract features that contain the appearance as well as the motion of the input video for anomaly detection. Many AE-based methods use explicit motion information with a pre-trained network such as an optical flow network~\cite{vu2019robust, liu2018future}, pose estimator \cite{markovitz2020graph}. However, explicitly separating information forces inductive bias into the network and makes it dependent on the pre-trained model. This can degrade network capacity due to the strong prior and cannot fully exploit end-to-end spatio-temporal representations.
	
	Therefore, we propose an implicit two-path AE (ITAE) that implicitly focuses on appearance and motion information. As  it is difficult to capture motion information using a single AE, we suggest a structure with two encoders and a single decoder, in which the two encoders capture relatively static and dynamic features, and the decoder learns to combine and reconstruct them together as original inputs. In contrast to other two-encoder and two-decoder structures or frameworks with pre-trained feature extractors, we simply add one encoder path and design ITAE with few shallow layers to make it suitable for video anomaly detection. Inspired by the SlowFast network \cite{feichtenhofer2019slowfast}, which achieves satisfactory performance in action recognition tasks, the static and dynamic encoders of ITAE have different temporal and channel sizes to focus either on appearance or on motion information. In Fig.~\ref{f1}, we visualize the output results when the appearance of the input frame looks normal (person) but the motion (riding a skateboard) is abnormal. Compared to one-path AE, the proposed ITAE gives a larger reconstruction error because the motion of the input frame differs from the learned normal frames. This large error leads to anomaly detection, which demonstrates that the dynamic encoder of ITAE is critical.
	
	For complex and diverse scenes, AE becomes difficult to reconstruct and has limitations in detecting anomalies. We suggest compensating for this drawback by distribution learning of latent features extracted from ITAE with a normalizing flow (NF)-based generative model. The method of classifying real reconstructed samples using a discriminator through adversarial learning~\cite{ravanbakhsh2017abnormal} faces limitations on supporting data distribution and is usually unstable because of  its formulation (min-max). In contrast, the NF model~\cite{dinh2014nice} focuses on the density estimation of high-dimensional data using tractable exact log-likelihood. By directly maximizing the likelihood of the NF model, it is possible to learn high-dimensional normal video features. For distribution modeling, instead of the original input frame, we use latent features from the proposed ITAE that represent normal patterns and show satisfactory results, especially in the ST database~\cite{liu2018future} with the most diverse anomalies. After training, abnormal events are found by out-of-distribution detection within ITAE feature likelihood (Fig.~\ref{f1}~(d)).
	The contributions of this paper are as follows:
	\begin{itemize}
		\item A  new approach, ITAE, is proposed, in which two encoders implicitly focus on static and dynamic features and a decoder learns to combine them to detect an event showing abnormal appearance or motion in surveillance videos.
		\item For complex distributions of normal scenes, using the features from ITAE, NF models estimate the density of normality and show the effectiveness of modeling learned using the ITAE embeddings. 
		\item Unsupervised learning without external datasets or models is conducted, and competitive performance on six surveillance anomaly detection benchmarks is achieved.
	\end{itemize}
	
	\section{Related Works}
	\subsection{Convolutional Networks for Video.} 
	As a basic network for video data, 3D convolution-based networks have been proposed and used for feature extraction in video anomaly detection. Recently, a transformer-based model for video representation learning has been studied~\cite{bertasius2021space, arnab2021vivit,fan2021multiscale,liu2021video}. TimeSformer~\cite{bertasius2021space} and Video Vision Transformer (ViViT)~\cite{arnab2021vivit} divide space and time information to perform self-attention using a sequence of spatio-temporal tokens; Multiscale Vision Transformers (MViT)~\cite{fan2021multiscale} contains several channel-resolution scale stages to create a multiscale pyramid of feature activation; Swin-T~\cite{liu2021video} extends the Swin Transformer to the spatio-temporal domain for spatio-temporal locality of videos. These methods usually utilize a Resnet backbone because of their low throughput. For video recognition, two-stream networks have also been proposed to model motion features explicitly. However, these require the explicit extraction of temporal information, such as temporal differences or optical flows. A Siamese network structure with contrastive learning~\cite{hadsell2006dimensionality} that maximizes the similarity of positive pairs while minimizing those of the negative ones has been studied as an unsupervised learning paradigm for representation learning. Furthermore, Multi-view Contrastive Learning with Noise-robust loss (MvCLN)~\cite{yang2021partially} has been proposed to learn consistent representations from multi-view/modal data.
	Recently, for video recognition, Feichtenhofer \etal~\cite{feichtenhofer2019slowfast} proposed slow and fast pathway networks that operate at different temporal and spatial resolutions to extract stationary and motion information. The slow-pathway is composed of a narrow temporal window, while the fast-pathway uses a higher temporal rate. In this paper, to learn normal appearance and motion patterns through reconstructing frames, we propose ITAE, which has two encoders and a single decoder composed of shallow layers, excluding residual blocks.
	
	\subsection{Video Anomaly Detection.} 
	Many anomaly detection algorithms based on frame reconstruction that harness the powerful representation ability of deep convolutional networks have been proposed. These algorithms exploit the structure of convolutional AEs~\cite{hasan2016learning}, recurrent neural networks~\cite{luo2017revisit}, or 3D convolutions~\cite{zhao2017spatio}. Other algorithms have been proposed by learning reconstruction with other objectives~\cite{nguyen2019hybrid}, usingmemory modules~\cite{gong2019memorizing, park2020learning}, reconstructing optical flows from frames~\cite{vu2019robust}, or encoding normal patterns with sparse dictionary learning~\cite{zhou2019anomalynet, luo2019video}. Zhou~\etal~\cite{zhou2019anomalynet} suggested a sparse long short-term memory (SLSTM) unit with adaptive ISTA $l_1$-solvers to encapsulate the historical information and achieve sparse codes in an unsupervised manner for anomaly detection; Luo~\etal~\cite{luo2019video} proposed a Temporally-coherent Sparse Coding (TSC) framework, which has a special type of sRNN that preserves the similarities between neighboring frames. Frame prediction-based approaches have been proposed to increase the probability of unpredictability for abnormal samples~\cite{liu2018future, nguyen2019anomaly, park2022fastano}. However, prediction-based approaches generally require heavier structures or optical flows, and bi-directional models rely heavily on future frames.
	
	As another approach, anomaly detection methods that learn the compactness of normal clusters has been proposed. Clusters are configured by removing small clusters of normal samples, features from the reconstruction objective~\cite{xu2015learning}, or extracting from a pre-trained object detector~\cite{ionescu2019object} or pose estimator~\cite{markovitz2020graph}. 
	
	\begin{figure*}[!t]
		\centering
		\includegraphics[width=0.9\linewidth]{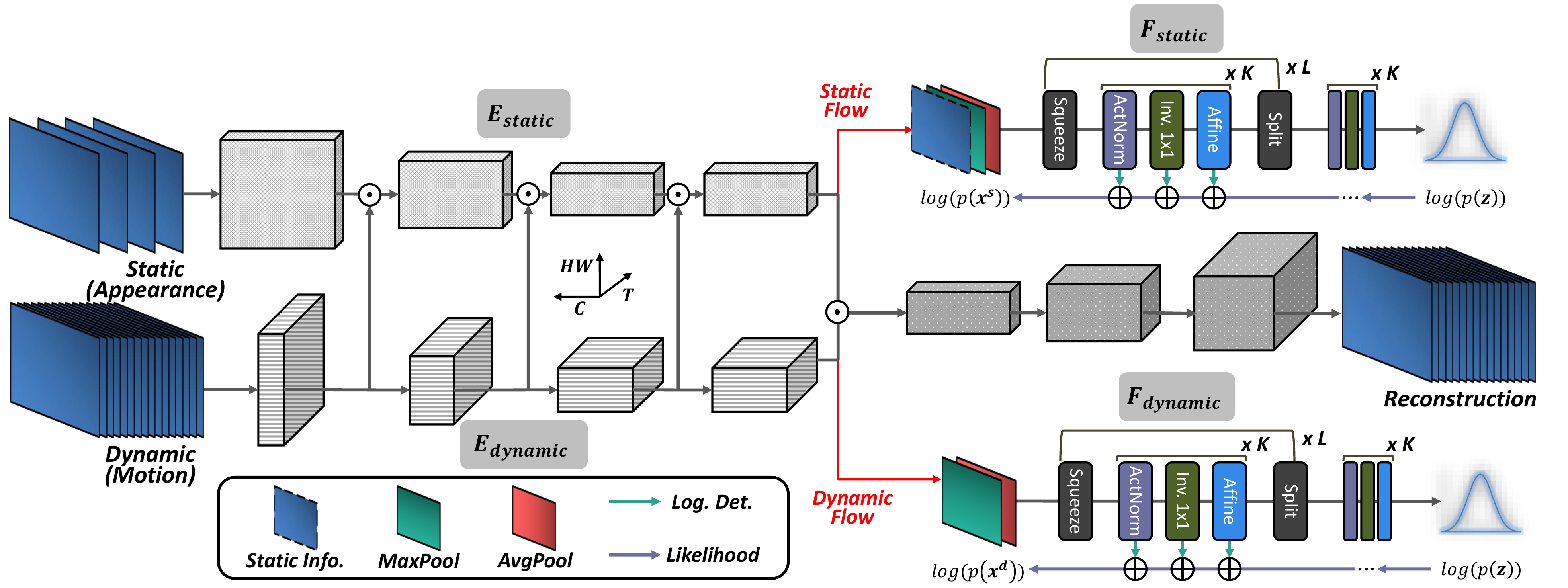}
		\caption{Overall framework of the proposed approach. Different frame rates are input into the static and dynamic encoders that focus on appearance and motion information, respectively. The latent features are concatenated and reconstructed into input frames by a single decoder. Furthermore, the max-pooling and average-pooling of each latent feature are concatenated and passed through the NF models to estimate the likelihood of normal static and dynamic features.}
		\label{framework}
	\end{figure*}
	
	\subsection{Generative Models.} 
	Compared with discriminative models, generative models do not require either normal annotations or proxy tasks such as frame generation. Some approaches based on a Gaussian model or non-parametric density estimation have been proposed based on either latent features or extracted features. 
	Deep generative models can be categorized into implicit and explicit density estimation models~\cite{goodfellow2016nips}. However, implicit density models, such as GAN-based anomaly detectors~\cite{wang2018anomaly}, do not define the data likelihood and cannot use in-distribution estimation without modifying the discriminator for the likelihood estimator, whereas explicit models first define likelihoods and try to maximize them. An approximation-based model (e.g., VAE) has been proposed to calculate likelihood, which requires approximation through a lower bound of the likelihood (ELBO), and has been used for anomaly detection. Utilizing Expectation Maximization (EM) to estimate density for anomaly detection has also been proposed~\cite{ouyang2021video}, which clusters features and applies E-step to retrieve the posterior likelihoods and M-step to update parameters iteratively. Auto-regressive models have shown promising results in estimating the density more precisely and have been adapted for anomaly detection~\cite{abati2019latent}. However, these autoregressive models are not efficient for high-dimensional data, cannot be implemented with parallel processing, and are sensitive to the choice of sequence order. 
	
	Tractable density estimators using NFs have been proposed to alleviate these issues~\cite{dinh2014nice, kingma2018glow}. Dinh \etal~\cite{dinh2014nice} proposed invertible networks using an affine coupling layer and calculated the tractable likelihood using the change of variable theorem. Kingma and Dhariwal proposed Glow~\cite{kingma2018glow} for further improvements using activation normalization and invertible $1\times1$ convolution. In this paper, we suggest distribution learning with NF models using static and dynamic features obtained from ITAE, which helps to deal with the complex distribution of normal scenes. Although there have been studies using likelihood for anomaly detection, the proposed method differs from them in the following aspects: (1) we compute exact tractable likelihood through normalizing flow, and (2) we perform density estimation of features learned with two path encoders and show effectiveness through experiment analysis, especially on the complex and multi-scene ShanghaiTech dataset, without using any pre-trained external models. To the best of our knowledge, this is the first attempt at learning appearance and motion normality through NFs in video anomaly detection, and it demonstrates effectiveness.
	
	\section{Proposed Method}
	\subsection{Overview}
	In surveillance anomaly detection, normal and anomalous scenes have differences in appearance (e.g., a car driving down a sidewalk), motion (e.g., jumping), or both (e.g., chasing a person with an abnormal object). Therefore, to learn the appearance and motion normal patterns of both scene types, we propose an ITAE that implicitly focuses on static and dynamic features. With these learned features from ITAE, we suggest distribution modeling through NF models to learn complex and diverse normality.
	
	\begin{table}[!t]
		\centering
		\caption{Instantiation of ITAE. Numbers in parentheses denote kernel size (temporal$\times$spatial). Output size is in the order of \{channel$\times$temporal$\times$spatial\} size.}
		\resizebox{0.9\linewidth}{!}{
			\renewcommand{\arraystretch}{1}
			{
				\begin{tabular}{c|c|c|c}
					\hline
					\multicolumn{1}{c|}{\multirow{2}[1]{*}{\textbf{Layer}}} & \multicolumn{2}{c|}{\textbf{Encoder}} & \multicolumn{1}{c}{\multirow{2}[1]{*}{\textbf{Output size}}} \\ \cline{2-3}
					\multicolumn{1}{c|}{} & \multicolumn{1}{c|}{\textbf{Static}} & \multicolumn{1}{c|}{\textbf{Dynamic}} & \multicolumn{1}{c}{} \\
					\hline
					\hline
					\multicolumn{1}{c|}{\multirow{2}[1]{*}{Conv1}} & \multicolumn{1}{c|}{($1, 3^2$)} & \multicolumn{1}{c|}{($5, 3^2$)} & \multicolumn{1}{c}{S: $96\times4\times128^2$} \\ 
					\multicolumn{1}{c|}{} & \multicolumn{1}{c|}{stride: $1, 2^2$} & \multicolumn{1}{c|}{stride: $1, 2^2$} & \multicolumn{1}{c}{D: $12\times16\times128^2$}\\
					\hline
					\multicolumn{1}{c|}{\multirow{2}[1]{*}{Conv2}} & \multicolumn{1}{c|}{($1, 3^2$)} & \multicolumn{1}{c|}{($3, 3^2$)} & \multicolumn{1}{c}{S: $128\times4\times64^2$} \\ 
					\multicolumn{1}{c|}{} & \multicolumn{1}{c|}{stride: $1, 2^2$} & \multicolumn{1}{c|}{stride: $1, 2^2$} & \multicolumn{1}{c}{D: $16\times16\times64^2$}\\
					\hline
					\multicolumn{1}{c|}{\multirow{2}[1]{*}{Conv3}} & \multicolumn{1}{c|}{($3, 3^2$)} & \multicolumn{1}{c|}{($3, 3^2$)} & \multicolumn{1}{c}{S: $256\times4\times64^2$} \\ 
					\multicolumn{1}{c|}{} & \multicolumn{1}{c|}{stride: $1, 1^2$} & \multicolumn{1}{c|}{stride: $1, 1^2$} & \multicolumn{1}{c}{D: $32\times16\times64^2$}\\
					\hline
					\multicolumn{1}{c|}{\multirow{2}[1]{*}{Conv4}} & \multicolumn{1}{c|}{($3, 3^2$)} & \multicolumn{1}{c|}{($3, 3^2$)} & \multicolumn{1}{c}{S: $256\times4\times64^2$} \\ 
					\multicolumn{1}{c|}{} & \multicolumn{1}{c|}{stride: $1, 1^2$} & \multicolumn{1}{c|}{stride: $1, 1^2$} & \multicolumn{1}{c}{ D: $32\times16\times64^2$}\\
					\hline
					\multicolumn{1}{c|}{\textbf{Layer}}&\multicolumn{2}{c|}{\textbf{Decoder}}&\multicolumn{1}{c}{\textbf{Output size}} \\
					\hline
					\hline
					\multicolumn{1}{c|}{\multirow{2}[1]{*}{DeConv1}} & \multicolumn{2}{c|}{($3, 3^2$)} & \multicolumn{1}{c}{\multirow{2}[1]{*}{$256\times4\times64^2$}} \\
					\multicolumn{1}{c|}{}&\multicolumn{2}{c|}{stride: $1, 1^2$}& \multicolumn{1}{c}{}\\
					\hline
					\multicolumn{1}{c|}{\multirow{2}[1]{*}{DeConv2}} & \multicolumn{2}{c|}{($3, 3^2$)} & \multicolumn{1}{c}{\multirow{2}[1]{*}{$128\times8\times128^2$}} \\
					\multicolumn{1}{c|}{}&\multicolumn{2}{c|}{stride: $2, 2^2$}& \multicolumn{1}{c}{}\\
					\hline
					\multicolumn{1}{c|}{\multirow{2}[1]{*}{DeConv3}} & \multicolumn{2}{c|}{($3, 3^2$)} & \multicolumn{1}{c}{\multirow{2}[1]{*}{$96\times16\times256^2$}} \\
					\multicolumn{1}{c|}{}&\multicolumn{2}{c|}{stride: $2, 2^2$}& \multicolumn{1}{c}{}\\
					\hline
					\multicolumn{1}{c|}{\multirow{2}[1]{*}{DeConv4}} & \multicolumn{2}{c|}{($3, 3^2$)} & \multicolumn{1}{c}{\multirow{2}[1]{*}{$3\times16\times256^2$}} \\
					\multicolumn{1}{c|}{}&\multicolumn{2}{c|}{stride: $1, 1^2$}& \multicolumn{1}{c}{}\\
					\hline
				\end{tabular}
			}
		}
		\label{t1}
	\end{table}
	
	The framework is trained in two steps using only normal training videos. In the first step, ITAE learns to reconstruct the normal frames. As depicted in Fig.~\ref{framework}, the sequence of frames is embedded through the ITAE with the static encoder ($E_{static}$) and dynamic encoder ($E_{dynamic}$). The embedding features of the two encoders are combined and reconstructed into original frames through a single decoder. In the second step, to estimate the density of the normal appearance and motion pattern, NF models ($F_{static}$ and $F_{dynamic}$) learn the distribution modeling of the two embedding features from the ITAE. In this step, ITAE is frozen to prevent the NF model from becoming difficult to be optimized when the weight of ITAE is updated and the feature distribution is changed.
	
	During testing, the abnormality score is calculated using the reconstruction error of the ITAE and the estimated likelihood of the NF models. When an abnormal scene is input, the ITAE learned using normal frames outputs a large error because of its poor reconstruction. Furthermore, with NF models, the static and dynamic embedding features obtained from ITAE differ from the features of the normal training set, which entails a low likelihood value.
	
	\subsection{Implicit two-path AE (ITAE)}
	\noindent
	\textbf{Two-path encoder.} 
	We design the AE as a two-path encoder to capture appearance and motion information implicitly and mutually. Surveillance videos tend to have static information that shows minimal change, such as background between frames, whereas dynamic information such as walking or running changes relatively quickly. The input sequences of the two encoders are sampled at different frame rates, and information is transferred between the two encoders through a lateral connection. We input $T/\tau$ frames at $\tau$ sampling rate to the static encoder, and input $T$ frames to the dynamic encoder (we use $\tau = 4$ in this study). A lateral connection concatenates the static and dynamic features by matching the temporal size with a $(5, 1, 1)$ kernel. 
	
	As the AE gets deeper and the parameter increases, the capacity becomes too powerful to reconstruct the anomalies accurately, so we compose each path into four shallow layers. In Table~\ref{t1}, each encoder composed with different spatial and temporal sizes of kernels and output features is presented. Furthermore, instead of a framework consisting of two encoder-decoder networks, ITAE fuses the features of two encoders into one latent feature, which the decoder uses to generate the original frame. These two encoders generate a higher reconstruction error than a one-path encoder for scenes with abnormal motion or appearance and perform better anomaly detection (visualized error maps are depicted in Fig.~\ref{f3} in Section \ref{visualization}).
	
	\noindent
	\textbf{Decoder.}
	The two embedding features obtained from each encoder are reconstructed into original input frames through a decoder. Similar to the lateral connection between encoders, the two final features are channel-wise concatenated and reconstructed (Fig.~\ref{framework}). For the decoder to generate the output by grasping the relationships between the static and dynamic features, we do not use the skip connection between the encoder and decoder, which is used primarily in the U-Net AE structure. Through encoders that focus on static and dynamic changes of input scenes and a decoder that fuses and reconstructs them by combining two embedded features, the network has the capacity to examine both sources of information implicitly. Without a complex structure, the ITAE that consists of four layers in each encoder and decoder is more powerful than the other proposed AE, even when using inception blocks or a convLSTM structure with a pre-trained network.
	
	\subsection{Learning Normality Distribution}
	We can learn normality from unlabeled normal training data by the unsupervised density estimation method. By using explicit likelihood generative models, it is possible to compute the likelihood of input data. An NF-based generative model can calculate the tractable likelihood via changes of variables toward a simple distribution (e.g., multivariate Gaussian). The likelihood is calculated by passing an invertible parametric function composed of multiple layers that maps the complex data distribution to the simple distribution, which can be any distribution. In this study, we set the prior $\boldsymbol{p_z}$ to be an isotropic unit norm Gaussian. With an input variable $\boldsymbol{x}\in X$, the distribution of which is unknown, a simple distribution $\boldsymbol{z}\sim p_{z}$, and a parametric function $\boldsymbol{f}_{\boldsymbol{\theta}}:X\rightarrow Z$, the integral of the probability density function is
	\begin{equation}
	\begin{split}
	\int _{\boldsymbol{z}}p_{z}(\boldsymbol{z}) \, \textrm{d}\boldsymbol{z} &= \int _{\boldsymbol{x}}p_{x}(\boldsymbol{x}) \, \textrm{d}\boldsymbol{x} \\ 
	&= \int _{\boldsymbol{x}}p_{z}(\boldsymbol{f}_{\boldsymbol{\theta}}(\boldsymbol{x}))\left | \textrm{det}\left (  \frac{\partial \boldsymbol{f}_{\boldsymbol{\theta}}}{\partial \boldsymbol{x}}\right ) \right | \, \textrm{d}\boldsymbol{x},
	\label{e1}
	\end{split}
	\end{equation}
	where $\textrm{det}\left ( \frac{\partial \boldsymbol{f}_{\boldsymbol{\theta}}}{\partial \boldsymbol{x}}\right )$ is the Jacobian determinant of function $\boldsymbol{f}_{\boldsymbol{\theta}}$ under change of variable theorem. When the generative model $\boldsymbol{f}$ with parameter $\boldsymbol{\theta}$ is $\boldsymbol{f}_{\boldsymbol{\theta}}=\boldsymbol{f}_{\boldsymbol{\theta_{M}}}\circ \boldsymbol{f}_{\boldsymbol{\theta_{M-1}}} \circ \cdots  \circ \boldsymbol{f}_{\boldsymbol{\theta_{1}}}$ and  $\boldsymbol{h}_{i}=\boldsymbol{f}_{\theta_{i}}(\boldsymbol{h}_{i-1})$ with $\boldsymbol{h}_{i-1} \sim p_{i-1}(\boldsymbol{h}_{i-1})$ where $\boldsymbol{h}_{0}=\boldsymbol{x}$ and $\boldsymbol{h}_{M}=\boldsymbol{z}$, the probability density function $p_{i}$ is as follows (for brevity, we omit $\boldsymbol{\theta}$ from $\boldsymbol{f_{\boldsymbol{\theta}}}$):
	
	\begin{align}
	p_i(\boldsymbol{h}_i) &= p_{i-1}(\boldsymbol{f}^{-1}_{i}(\boldsymbol{h}_i)) \left | \textrm{det}\left ( \frac{\partial \boldsymbol{f}^{-1}_{i}}{\partial \boldsymbol{h}_{i}} \right ) \right |\label{eq1-3-2} \\ 
	&= p_{i-1}(\boldsymbol{h}_{i-1}) \left | \textrm{det}\left ( \frac{\partial \boldsymbol{f}_{i}}{\partial \boldsymbol{h}_{i-1}} \right )^{-1} \right |\label{eq1-3-3}\\
	&= p_{i-1}(\boldsymbol{h}_{i-1}) \left | \textrm{det}\left ( \frac{\partial \boldsymbol{f}_{i}}{\partial \boldsymbol{h}_{i-1}} \right ) \right |^{-1}\label{eq1-3-4}
	\end{align}
	
	Equation~\ref{eq1-3-2} is written as Eq.~\ref{eq1-3-3} according to the inverse function theorem: if $y = f(x)$ and $x = f^{-1}(y)$ then $\frac{\textsl{d}f^{-1}(y)}{\textsl{d}y} = \frac{\textsl{d}x}{\textsl{d}y} = (\frac{\textsl{d}y}{\textsl{d}x})^{-1} = (\frac{\textsl{d}f(x)}{\textsl{d}x})^{-1}$.  Equation~\ref{eq1-3-4} is from Jacobians of invertible function of $\textrm{det}(M^{-1}) = (\textrm{det}(M))^{-1}$.
	Then,
	\begin{align}
	\textsl{log}\, p_i(\boldsymbol{h}_i) = \textsl{log}\, p_{i-1}(\boldsymbol{h}_{i-1}) - \textsl{log} \left | \textrm{det}\left ( \frac{\partial \boldsymbol{f}_{i}}{\partial \boldsymbol{h}_{i-1}} \right ) \right |\label{eq1-3-5}
	\end{align}
	By repeatedly applying the rule for change of variables, the expanded equations are as follows:
	\begin{gather}
	\textsl{log}\, p_z(\boldsymbol{z}) = \textsl{log}\, p_{M-1}(\boldsymbol{h}_{M-1}) - \textsl{log} \left | \textrm{det}\left ( \frac{\partial \boldsymbol{f}_{M}}{\partial \boldsymbol{h}_{M-1}} \right ) \right |\nonumber \\
	= \textsl{log}\, p_{M-1}(\boldsymbol{h}_{M-1}) - \textsl{log} \left | \textrm{det}\left ( \frac{\partial \boldsymbol{f}_{M-1}}{\partial \boldsymbol{h}_{M-2}} \right ) \right | - \textsl{log} \left | \textrm{det}\left ( \frac{\partial \boldsymbol{f}_{M}}{\partial \boldsymbol{h}_{M-1}} \right ) \right |\nonumber \\
	\vdots\nonumber\\
	= \textsl{log}\, p_{x}(\boldsymbol{x}) - \textsl{log} \left | \textrm{det}\left ( \frac{\partial \boldsymbol{f}_{1}}{\partial \boldsymbol{x}} \right ) \right | - \cdots - \textsl{log} \left | \textrm{det}\left ( \frac{\partial \boldsymbol{f}_{M}}{\partial \boldsymbol{h}_{M-1}} \right ) \right |\nonumber \\
	= \textsl{log}\, p_{x}(\boldsymbol{x}) - \sum^{M}_{i=1} \textsl{log} \left | \textrm{det}\left ( \frac{\partial \boldsymbol{f}_{i}}{\partial \boldsymbol{h}_{i-1}} \right ) \right |
	\label{eq1-3-6}
	\end{gather}
	With these steps, the probability density function of $\boldsymbol{x}$ is as given in Eq.~\ref{e2}.
	\begin{gather}
	log\, p_{x}(\boldsymbol{x}) = log\, p_{z}(\boldsymbol{z}) + \sum^{M}_{i=1}log\,\left | \textrm{det}\left ( \frac{\partial \boldsymbol{f}_{i}}{\partial \boldsymbol{h}_{i-1}} \right ) \right |
	\label{e2}
	\end{gather}
	If the density function of the latent variable $\boldsymbol{z}$ is tractable like a Gaussian distribution and the Jacobian matrix $\frac{\partial \boldsymbol{f}_{\boldsymbol{\theta}}}{\partial \boldsymbol{x}}$ is triangular, the likelihood of input variable $\boldsymbol{x}$ can be obtained simply. We use the Glow model with $L$ level multi-scale architecture and $K$ series of Actnorm, invertible convolution, and Affine layer for density estimation~\cite{kingma2018glow}.
	
	By maximizing the likelihood in Eq.~\ref{e2}, an NF model estimates the density of high-dimensional data through multiple layers of convolutional networks. As the likelihood of a generative model heavily depends on the image complexity~\cite{serra2019input}, unlike Glow, which begins from the image space, the intermediate latent feature of ITAE is used for complex density modeling of normal videos. After training the ITAE with frame reconstruction, we estimate the density of the static and dynamic embedding features obtained from each encoder. The max-pooling and average-pooling along the channel axis of the feature from each path are applied to reduce the dimensions and are concatenated and input into each NF model, i.e., $F_{static}$ and $F_{dynamic}$ (Fig.~\ref{framework}). We also concatenate the resized intensity of the input frame to the $F_{static}$ input, which provides additional sparse appearance information of the feature map. For abnormal scenes, static and dynamic NF models trained with latent features that embed each appearance and motion pattern of normal scenes show low likelihood results. 
	
	\subsection{Training and Testing}
	\noindent\textbf{Reconstruction loss function.}
	For reconstruction, the model is trained by minimizing the L2 loss of the input sequence of frames $\boldsymbol{I}$ (the ground truth) and the output frames $\boldsymbol{\hat I}$ to make all pixels of the RGB or gray channel similar (Eq.~\ref{e3_1}). We also add multi-scale SSIM $L_{ms-ssim}$ and gradient loss $L_{grad}$ to maintain the sharpness of the frames, which computes the difference of gradient at each pixel between the input and output frames. The total loss is the sum of $L_2$, $L_{ms-ssim}$, and $L_{grad}$ as given in Eq.~\ref{e3}. Just as~\cite{liu2018future, nguyen2019anomaly, tang2020integrating} used the sum of the intensity loss and gradient loss having the same importance as per~\cite{mathieu2015deep}, we also add the three loss terms having the same importance by following~\cite{lu2019future}.
	
	\begin{gather}\label{e3_1}
	L_{2} = \left \| \boldsymbol{I}-\boldsymbol{\hat{I}} \right \|^{2}_{2}
	\end{gather}
	\begin{gather}
	\hspace{0.7em} L_{recon} = L_{2} + L_{ms-ssim} + L_{grd}
	\label{e3}
	\end{gather}
	
	\noindent
	\textbf{Log-likelihood loss function.}
	After training ITAE, the generative models $F_{static}$ and $F_{dynamic}$ are trained with the negative log-likelihood (nll) $L_{nll}$ of the static and dynamic embedding feature $\boldsymbol{x}^s$ and $\boldsymbol{x}^d$ in Eq.~\ref{e7}. As in Eq.~\ref{e2}, the exact log-likelihood $log\, p_{x}(\boldsymbol{x}; \boldsymbol{\theta})$ of the input feature is calculated through generative models, and the parameters $\boldsymbol{\theta}$ are updated to minimize nll in Eq.~\ref{e6}.
	\begin{gather}
	L_{nll} = NLL(\boldsymbol{x}^{s}) + NLL(\boldsymbol{x}^{d})
	\label{e7}
	\end{gather}
	\begin{gather}
	\boldsymbol{\theta}^{*} = \argmin_\theta\; - log\,p_{x}(\boldsymbol{x};\boldsymbol{\theta})\nonumber \\ 
	NLL(\boldsymbol{x}) = -log\,p_{X}(\boldsymbol{x})
	\label{e6}
	\end{gather}
	
	\noindent
	\textbf{Anomaly score.}
	The anomaly score for the reconstruction error $R(\boldsymbol{I}_{t}, \boldsymbol{\hat I}_{t})$ of the $t$-th frame is the difference between input and the output frame of the ITAE within a sliding patch (in Eq.~\ref{e8}, $P$ indicates an $N \times N$ image patch and $|P|$ is the pixel number of the patch). For each frame, we compute the mean of error values in all segments in which it appears. The score $L(\boldsymbol{x}_{t}^{s}, \boldsymbol{x}_{t}^{d})$ from the generative models is calculated by adding the nll values of each static feature and the dynamic feature and normalizing it in Eq.~\ref{e9} where $\textrm{norm}(\cdot)$ denotes normalization within a video clip as in some previous studies~\cite{liu2018future, abati2019latent, park2020learning}.
	
	\begin{gather}
	R(\boldsymbol{I}_{t}, \boldsymbol{\hat I}_{t}) = \argmax_{sliding P}\frac{1}{|p|}\sum_{i,j\in P}| \boldsymbol{I}_{t}^{i,j}-\boldsymbol{\hat I}_{t}^{i,j} |
	\label{e8}
	\end{gather}
	\begin{gather}
	L(\boldsymbol{x}_{t}^{s}, \boldsymbol{x}_{t}^{d}) = norm(NLL(\boldsymbol{x}_{t}^{s}) + NLL(\boldsymbol{x}_{t}^{d}))
	\label{e9}
	\end{gather}
	The total anomaly score $S_{t}$ is computed by summing the reconstruction error and nll with scaling factor $\lambda_L$ in Eq.~\ref{e10}. $\lambda_L$ is experimentally obtained value from the set [0.1, 0.3, 0.5, 0.7, 0.9, 1.0] (in Section~\ref{lambda}).
	\begin{gather}
	S_{t} = R(\boldsymbol{I}_{t}, \boldsymbol{\hat I}_{t})+\lambda_{L}L(\boldsymbol{x}_{t}^{s}, \boldsymbol{x}_{t}^{d})
	\label{e10}
	\end{gather}
	
	\section{Experiments}
	For the evaluation, we compute the average area under curve (AUC) and Equal Error Rate (EER) through the receiver operation characteristic (ROC) by gradually changing the threshold of the anomaly score in a frame-level annotated database (the ROC curves are reported in the supplementary material). For a fair comparison, we report Mirco-AUC performance including \cite{georgescu2020background, ionescu2019object, georgescu2021anomaly}.
	
	\subsection{Databases}
	\noindent
	\textbf{UCSD \cite{li2013anomaly}.}
	UCSD consists of two subsets, Ped1 and Ped2, which are composed of overlooking walkway scenes obtained through a mounted camera. The foreground object size and movement changes are small, and videos are grayscale and low-resolution. Ped1 has a very low frame resolution, with an inconsistency issue; for example, some \cite{ravanbakhsh2017abnormal, xu2015learning} tested on only 16 videos, while others \cite{hasan2016learning, zhao2017spatio} tested all videos. For these reasons, we conduct experiments only on Ped2 as in another study \cite{nguyen2019anomaly, tang2020integrating}. There are anomalies of non-pedestrian objects in the walkways, such as cars and skaters in the test set, and the density of pedestrians varies.
	
	\noindent
	\textbf{CUHK Avenue \cite{lu2013abnormal}.} CUHK consists of normal videos in the training set with several outliers included and a slight camera shake. The test set includes anomalies such as a person walking in the wrong direction, strange actions, substantial motion, and foreground scale variation. 
	
	\noindent
	\textbf{Shanghai Tech Campus (ST) \cite{liu2018future}.}
	ST is the largest-volume database, containing 13 different scenes, whereas the two databases above contain a single  scene. This database includes diverse anomalous events such as brawling and loitering, including sudden motion in multiple scenes. It is challenging because of its complex angles and lighting conditions.
	
	\noindent
	\textbf{UCF-Crimes~\cite{sultani2018real}.} UCF-Crimes consists of long untrimmed surveillance videos that cover 13 real-world anomalies, including abuse, arrest, arson, assault, accident, burglary, explosion, fighting, robbery, shooting, stealing, shoplifting, and vandalism. This is a large-scale anomaly detection dataset composed of 950 unedited real-world surveillance videos with clear anomalies and 950 normal videos, leading to a total of 1,900 videos with a total duration of 128 hours. Unlike an unsupervised anomaly detection dataset whose training set only consists of normal videos, it contains 1,610 training videos with video-level labels and 290 test videos with frame-level labels of much more complicated and diverse backgrounds.
	
	\noindent
	\textbf{Live-Videos (LV)~\cite{leyva2017lv}.} LV consists of 28 surveillance videos with a total duration of 3.93 hours captured at different frame rates in indoor and outdoor scenarios with several illumination changes and some camera motions. The anomalies include panic, robberies, kidnap, fighting, and clashes.
	
	\noindent
	\textbf{UBI-Fights~\cite{degardin2020human}.} UBI-Fight is a large-scale  abnormal event dataset that consists of 1,000 videos with a duration of 80 hours. There are 216 videos containing a fight event and 784 normal daily life videos captured under various conditions such as indoor, outdoor, grayscale, RGB-scale, rotating scenario, and moving camera.
	
	\subsection{Implementations}
	We consider three databases in our experiments: UCSD, CUHK, and ShanghaiTech (ST). For training, we resize the input frame to $256\times 256$ and set $T$ to 16. For UCSD, which has a small foreground scale, we use the original frame size ($240\times360$). We use the Adam optimizer and Cosine annealing scheduler. In the first training step, the batch sizes are 2, 2, and 8, and the learning rates are $1e^{-3}, 1e^{-2}$, and $1e^{-2}$ for the UCSD, CUHK, and ST databases, respectively. For the second step, the batch sizes are 8, 5, and 8, and the learning rates are $5e^{-4}, 5e^{-4}$, and $1e^{-4}$. $\lambda_{L}$ is 0.3, 0.1, and 0.7, respectively. Glow is used for the NF models with $K=32$ and $L=3$ ($L=1$ for UCSD). For the ST dataset, which contains multi-scene videos, each anomaly score is normalized to match the scale of the score. For real-world scenario datasets, UCF-Crimes, LV, and UBI-Fights, all settings are the same as ST. Please refer to the supplementary material for detailed information on experiments.
	
	\subsection{Ablation Studies}
	\subsubsection{Structure of ITAE}
	\begin{table}[!t]
		\centering
		\caption{Ablation studies of two-path AE on CUHK database.}
		\resizebox{1.0\linewidth}{!}{
			{\Large
				\begin{tabular}{ccccccc}
					\hline
					\textbf{AE} &
					\multicolumn{2}{|c}{\textbf{SlowFast-18 AE}} & \multicolumn{2}{|c|}{\textbf{SlowFast-50 AE}} & \multicolumn{2}{c}{\textbf{ITAE}} \\
					\cline{2-7} \textbf{Model} & \multicolumn{1}{|c}{\textbf{one path}} & \textbf{two path} & \multicolumn{1}{|c}{\textbf{one path}} & \textbf{two path} & \multicolumn{1}{|c}{\textbf{one path}} & \textbf{two path} \\
					\hline
					\hline
					\hline
					\multicolumn{1}{c|}{AUC} & 83.3 & \multicolumn{1}{c|}{81.1} & 81.8 & \multicolumn{1}{c|}{80.7}  & \textbf{86.4} & \textbf{87.3}\\
					\hline
			\end{tabular}}
		}
		\label{t0}
	\end{table}
	In Table~\ref{t0}, we compare the two-path networks SlowFast~\cite{feichtenhofer2019slowfast} and ITAE. In contrast to ITAE, SlowFast is a discriminative model proposed for action recognition that has deep ResNet-style layers. For comparison, we train SlowFast AE (Backbone as ResNet18 and 50) by attaching a decoder with residual layers to SlowFast (experiment conditions are the same as ITAE). As can be seen from the SlowFast-18 and 50 AE results, the performance of the two-path encoder is worse than when there is only a one-path encoder, which indicates that increased model complexity does not lead to performance gains. The deep and complex structure of SlowFast AE instead degrades the performance of detecting anomalies. In contrast, ITAE, with a shallow and non-residual network, performs better and improves markedly when both static and dynamic paths are considered together (Fig.~\ref{f3}). This result illustrates that ITAE is more suitable for video anomaly detection where motion and appearance information sources are important.
	
	\subsubsection{Distribution modeling with ITAE features}
	\begin{table}[!t]
		\centering
		\caption{Normal density estimation using raw frames and ITAE features with NF models on three benchmarks.}
		\resizebox{1.0\linewidth}{!}{
			\begin{threeparttable}
				\centering
				{\Large
					\begin{tabular}{c|>{\centering\arraybackslash}p{0.2\textwidth}|>{\centering\arraybackslash}p{0.2\textwidth}|>{\centering\arraybackslash}p{0.2\textwidth}}
						\hline
						\multirow{2}[1]{*}{\textbf{Database}} & \textbf{NF} & \multicolumn{2}{c}{\textbf{NF on ITAE features}} \\
						\cline{3-4}& \textbf{on Frames} &\textbf{ w/o ITAE }& \textbf{w ITAE} \\
						\hline
						\hline
						UCSD Ped2\tnote{$^{\star}$}~\cite{li2013anomaly} & 82.2 & 90.6 & \textbf{99.2} \\
						CUHK$^{\diamond}$~\cite{lu2013abnormal} & 79.4 & 74.6  & \textbf{88.0} \\
						Shanghai Tech$^{\dagger}$~\cite{liu2018future} & 69.2 & 72.5  & \textbf{76.3} \\
						\hline
					\end{tabular}
				}
				\begin{tablenotes}\footnotesize
					\item[] $^{\star}$ 1 scene/ 28 clips, \hspace{0.3em} $^{\diamond}$ 1 scene/ 38 clips, \hspace{0.3em} $^{\dagger}$ 13 scenes/ 437 clips
				\end{tablenotes}
			\end{threeparttable}
		}
		\label{t5}
	\end{table}
	
	\begin{figure*}[!t]
		\centering
		\subfloat[UCSD~\cite{li2013anomaly}]{\includegraphics[width=0.3\linewidth]{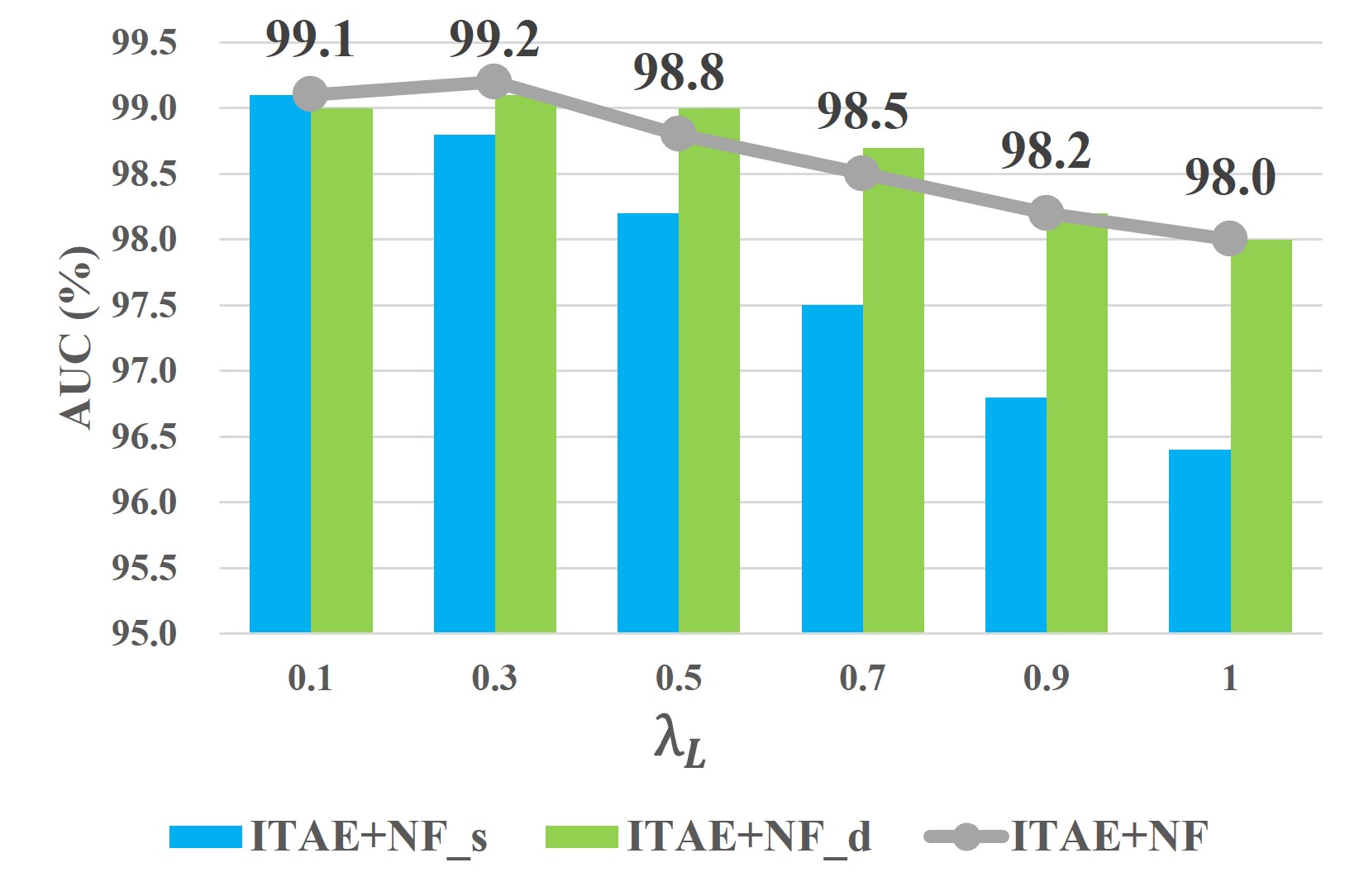}	\label{f4_3_3_a}}
		\subfloat[CUHK~\cite{lu2013abnormal}]{\includegraphics[width=0.3\linewidth]{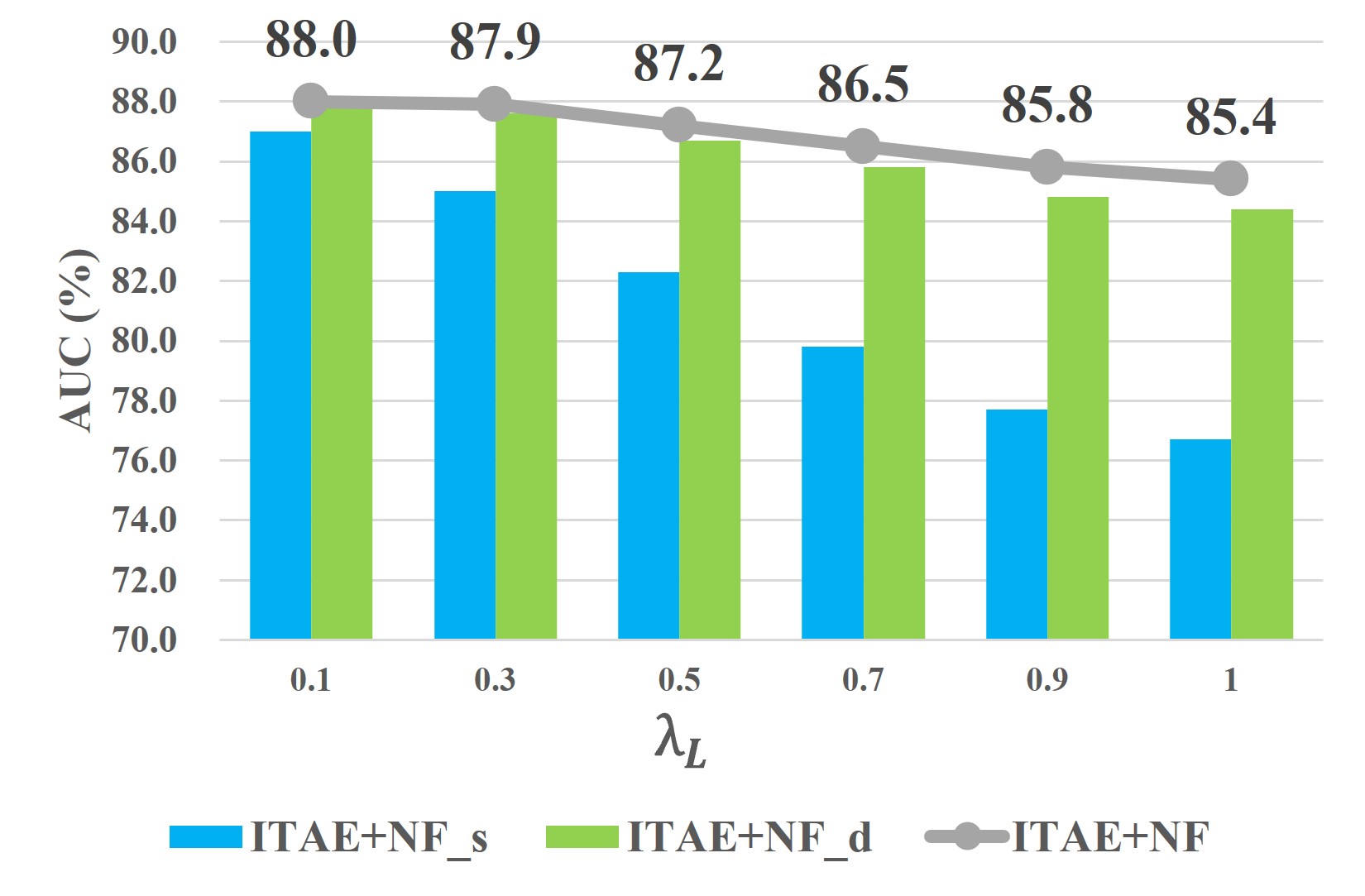}
			\label{f4_3_3_b}}
		\subfloat[ST~\cite{liu2018future}]{\includegraphics[width=0.3\linewidth]{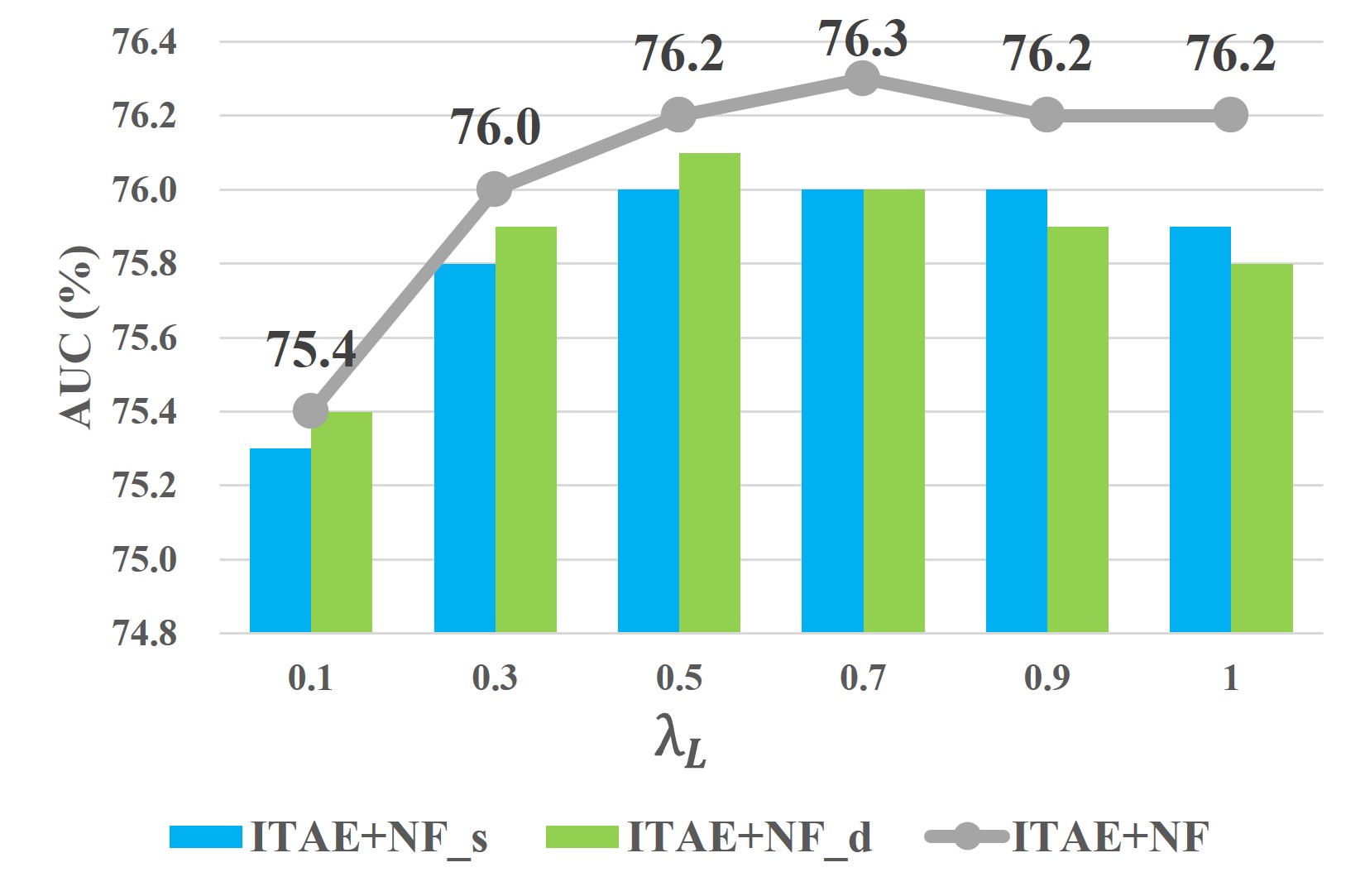}
			\label{f4_3_3_c}}
		\caption{AUC scores on three datasets obtained by selecting values for the $\lambda_{L}$ from the set [0.1, 0.3, 0.5, 0.7, 0.9, 1.0]}
		\label{f4_3_3}
	\end{figure*}
	
	Our framework estimates the normality using the ITAE features instead of the raw frame for distribution modeling. In Table~\ref{t5}, without an ITAE reconstruction score, NF models with ITAE static and dynamic features illustrate superior or comparable results to those with the raw frames ($128 \times 128$ input size), which proves the effectiveness of modeling with learned ITAE features for surveillance video data. Furthermore, in the ST dataset, i.e., the largest and most diverse among the three benchmarks (it has 13 scenes, whereas others have 1), the NF model exhibits excellent results without an ITAE score. With extensive and diverse data, NF performs more general distribution learning and responds to complex scenes.
	
	\subsubsection{Ablation studies of proposed framework} 
	\begin{table}[!t]
		\centering
		\caption{Ablation studies of static and dynamic encoders in ITAE and NF models on three benchmark databases.}
		\resizebox{1.0\linewidth}{!}{
			\begin{threeparttable}
				\renewcommand{\arraystretch}{1}
				{\Large
					\begin{tabular}{cccc|cc|cc|cc}
						\hline
						\multicolumn{2}{c|}{\textbf{ITAE}} & \multicolumn{2}{c|}{\textbf{NFs}}&
						\multicolumn{2}{c|}{\textbf{UCSD$^{\star}$}~\cite{li2013anomaly}} & \multicolumn{2}{c|}{\textbf{CUHK$^{\diamond}$}~\cite{lu2013abnormal}}&\multicolumn{2}{c}{\textbf{ST}$^{\dagger}$~\cite{liu2018future}} \\
						\cline{1-4} \multicolumn{1}{c}{\textbf{Static}} & \multicolumn{1}{c|}{\textbf{Dynamic}} & \multicolumn{1}{c}{\textbf{Static}} & \multicolumn{1}{c|}{\textbf{Dynamic}} & \textbf{AUC} & \textbf{EER }& \textbf{AUC} & \textbf{EER} & \textbf{AUC} & \textbf{EER}\\
						\hline
						\hline
						\checkmark & & & & 97.7 & 5.6 & 86.4 & 21.3 & 73.1 & 33.1\\
						\checkmark & \checkmark& & & 98.7 & 5.5 & 87.3 & 19.6 & 74.8 & 31.8\\
						\checkmark & \checkmark& \checkmark& & 98.8 & 4.4 &87.0 & 20.2 & 76.0 &\textbf{30.3}\\
						\checkmark &\checkmark & &\checkmark & 99.1 & 4.1&87.9&19.1&76.0&31.1\\
						\checkmark & \checkmark& \checkmark& \checkmark& \textbf{99.2}&\textbf{3.9}&\textbf{88.0}&\textbf{19.0}&\textbf{76.3}&30.6\\
						\hline
					\end{tabular}
					\begin{tablenotes}\footnotesize
						\item[] $^{\star}$ 1 scene/ 28 clips, \hspace{0.3em} $^{\diamond}$ 1 scene/ 38 clips, \hspace{0.3em} $^{\dagger}$ 13 scenes/ 437 clips
					\end{tablenotes}
				}
			\end{threeparttable}
		}
		\label{t3}
	\end{table}	 
	
	\begin{table}[!t]
		\centering
		\caption{Comparison with state-of-the-art methods on three benchmarks; the comparison result of our model is the performance of ITAE and static and dynamic NF models.}
		\resizebox{1.0\linewidth}{!}{
			\begin{threeparttable}
				\renewcommand{\arraystretch}{1}
				{\Large
					\begin{tabular}{c|c|c|cc|cc|cc}
						\hline
						\multirow{10}[25]{*}{\begin{sideways}\textbf{\large{w external dataset · model}}\end{sideways}} &
						\multicolumn{1}{c|}{\textbf{External}} &
						\multicolumn{1}{c|}{\multirow{2}[1]{*}{\textbf{Methods}}}& \multicolumn{2}{c|}{\textbf{UCSD$^{\star}$}~\cite{li2013anomaly}} & \multicolumn{2}{c|}{\textbf{CUHK$^{\diamond}$}~\cite{lu2013abnormal}} & \multicolumn{2}{c}{\textbf{ST$^{\dagger}$}~\cite{liu2018future}} \\
						\cline{4-9}&\multicolumn{1}{c|}{\textbf{module}}&\multicolumn{1}{c|}{}&\multicolumn{1}{c|}{\textbf{AUC}}&\multicolumn{1}{c|}{\textbf{ERR}}&\multicolumn{1}{c|}{\textbf{AUC}}&\multicolumn{1}{c|}{\textbf{ERR}}&\multicolumn{1}{c|}{\textbf{AUC}}&\multicolumn{1}{c}{\textbf{ERR}}\\
						\cline{2-9}
						&FRCN& MT-FRCN \cite{hinami2017joint}&92.2&13.9&-&-&-&-\\
						&OpticalFlow& AbnormalGAN \cite{ravanbakhsh2017abnormal}&93.5&14.0&-&-&-&-\\
						&Flownet& FFP~\cite{liu2018future}&95.4&-&84.9&-&72.8&-\\
						&Flownet2& AMC \cite{nguyen2019anomaly}&96.2&-&86.9&-&-&-\\
						&OpticalFlow& MLAD~\cite{vu2019robust}&97.5&\textbf{4.7}&71.5&36.4&-&-\\
						&Yolov3& Obj-centric~\cite{ionescu2019object}&94.3&-&87.4&-&\underline{78.7}&-\\
						&Resnet-50& AnomalyNet~\cite{zhou2019anomalynet}&94.9&10.2&86.1&\underline{22.0}&-&-\\
						&Flownet& UNet-inte~\cite{tang2020integrating}&96.3&\underline{10.0}&85.1&-&73.0&-\\
						&OpticalFlow& STFF~\cite{wu2020fast}&92.8&12.5&85.5&\textbf{20.7}&-&-\\
						&SelFlow, Yolov3& BG-Agnostic~\cite{georgescu2020background}&\textbf{98.7}&-&\textbf{92.3}&-&\textbf{82.7}&-\\
						&Flownet, Yolov3& MONAD~\cite{doshi2021online}&97.2&-&86.4&-&70.9&-\\
						&Yolov3& Multi-task~\cite{georgescu2021anomaly}&\underline{97.5}&-&\underline{91.5}&-&\underline{82.4}&-\\
						\hline
						\hline
						\multirow{15}[30]{*}{\begin{sideways}\textbf{\large{w/o external dataset/model}}\end{sideways}} &
						\multicolumn{2}{c|}{\multirow{2}[1]{*}{\textbf{Methods}}}& \multicolumn{2}{c|}{\textbf{UCSD$^{\star}$}~\cite{li2013anomaly}} & \multicolumn{2}{c|}{\textbf{CUHK$^{\diamond}$}~\cite{lu2013abnormal}} & \multicolumn{2}{c}{\textbf{ST$^{\dagger}$}~\cite{liu2018future}} \\
						\cline{4-9}&\multicolumn{2}{c|}{}&\multicolumn{1}{c|}{\textbf{AUC}}&\multicolumn{1}{c|}{\textbf{ERR}}&\multicolumn{1}{c|}{\textbf{AUC}}&\multicolumn{1}{c|}{\textbf{ERR}}&\multicolumn{1}{c|}{\textbf{AUC}}&\multicolumn{1}{c}{\textbf{ERR}}\\
						\cline{2-9}
						&& \multicolumn{1}{c|}{ AMDN \cite{xu2015learning}}&90.8&17.0&-&-&-&-\\
						&& \multicolumn{1}{c|}{BMAN~\cite{lee2019bman}}&96.6&-&\textbf{90.0}&-&76.2&-\\
						&\textbf{None-Recon.}& \multicolumn{1}{c|}{DDGAN~\cite{dong2020dual}}&95.6&-&84.9&-&73.7&32.0\\
						&& \multicolumn{1}{c|}{Mem-guided~\cite{park2020learning}}&97.0&-&88.5&-&70.5&-\\
						&& \multicolumn{1}{c|}{Fewshot~\cite{lu2020few}}&96.2&-&85.8&-&\textbf{77.9}&-\\
						&& \multicolumn{1}{c|}{STCEN~\cite{hao2022spatiotemporal}}&96.9&8.0&86.6&\underline{19.2}&73.8&-\\
						\cline{2-9}&& \multicolumn{1}{c|}{AE-Conv2D \cite{hasan2016learning}}&90.0&21.7&70.2&25.1&-&-\\
						&& \multicolumn{1}{c|}{AE-Conv3D \cite{zhao2017spatio}}&88.6&20.9&80.9&24.4&-&-\\
						&& \multicolumn{1}{c|}{TSC \cite{luo2017revisit}}&91.0&-&80.6&-&67.9&-\\
						&& \multicolumn{1}{c|}{StackRNN \cite{luo2017revisit} }&92.2&-&81.7&-&68.0&-\\
						&& \multicolumn{1}{c|}{HybridAE~\cite{nguyen2019hybrid} }&84.3&-&82.8&-&-&-\\
						&\textbf{Recon.}& \multicolumn{1}{c|}{Auto-reg~\cite{abati2019latent} }&95.4&-&-&-&72.5&-\\
						&& \multicolumn{1}{c|}{MemAE~\cite{gong2019memorizing}}&94.1&-&83.3&-&71.2&-\\
						&& \multicolumn{1}{c|}{LRCCDL~\cite{li2020abnormal}}&96.6&8.9&-&-&-&-\\
						&& \multicolumn{1}{c|}{Mem-guided~\cite{park2020learning}}&90.2&-&82.8&-&69.8&-\\
						&& \multicolumn{1}{c|}{DissociateAE~\cite{chang2022video}}&96.7&-&87.1&-&73.7&-\\
						&& \multicolumn{1}{c|}{\textbf{ITAE(ours)}} &\underline{98.7}&\underline{5.5}&87.3&19.6&74.8&\underline{31.8}\\
						&&\multicolumn{1}{c|}{\textbf{ITAE+NFs(ours)}} &\textbf{99.2}&\textbf{3.9}&\underline{88.0}&\textbf{19.0}&\underline{76.3}&\textbf{30.6}\\
						\hline
					\end{tabular}
					
					\begin{tablenotes}\footnotesize
						\item[] $^{\star}$ 1 scene/ 28 clips, \hspace{0.3em} $^{\diamond}$ 1 scene/ 38 clips, \hspace{0.3em} $^{\dagger}$ 13 scenes/ 437 clips
					\end{tablenotes}
				}
			\end{threeparttable}
			\label{t2}
		}
	\end{table}
	
	Table~\ref{t3} presents the results of ablation studies with each static and dynamic path of ITAE and NF models. Using the dynamic encoder produces superior results in all three databases compared with using only the static encoder. On the ST database, which contains the most complex and largest-volume videos, the most significant performance improvement, i.e., 1.7\%, is achieved when the dynamic path is added. Furthermore, with NF models, the normality distribution modeling of the ITAE latent feature shows the best performance. On CUHK, dynamic feature modeling presents more effective results than the static feature because anomalous events (throwing a bag or papers, running person on the walkway) are associated primarily with  motion. On the ST database, which has the largest number of training data sets and various anomalous events in the test set, the performance improves the most when the NF models added to ITAE–flow models learn more general distribution with a large amount of training~data.
	
	\subsubsection{Ablation studies of $\lambda_L$} \label{lambda}
	Figure~\ref{f4_3_3} shows the results of AUC and ERR on three benchmarks according to the hyper-parameter $\lambda_L$. In Eq.~\ref{e10}, the anomaly score $S_t$ is calculated as the sum of the reconstruction score $R(\boldsymbol{I}_{t}, \boldsymbol{\hat I}_{t})$ and the NF score $L(\boldsymbol{x}_{t}^{s}, \boldsymbol{x}_{t}^{d})$, and $\lambda_L$ is the scaling factor of the NF score. We compute the scores of $ITAE$ with static NF model $NF_{static}$, dynamic NF model $NF_{dynamic}$, and both NF models $NF$ by scaling $\lambda_L$ from the set [0.1, 0.3, 0.5, 0.7, 0.9, 1.0]. For UCSD, CUHK, and ST, we select the $\lambda_L$ value as 0.3, 0.1, and 0.7, respectively. In particular, in the ST dataset composed of multiple and diverse scenes, the highest performance improvement with highest importance of $\lambda_L=0.7$ is achieved, which indicates the effectiveness of the density model on diverse and complex scenes. (Please refer to the supplementary material for detailed results.)
	
	\begin{figure}[!t]
		\centering
		\subfloat{\includegraphics[width=0.28\columnwidth]{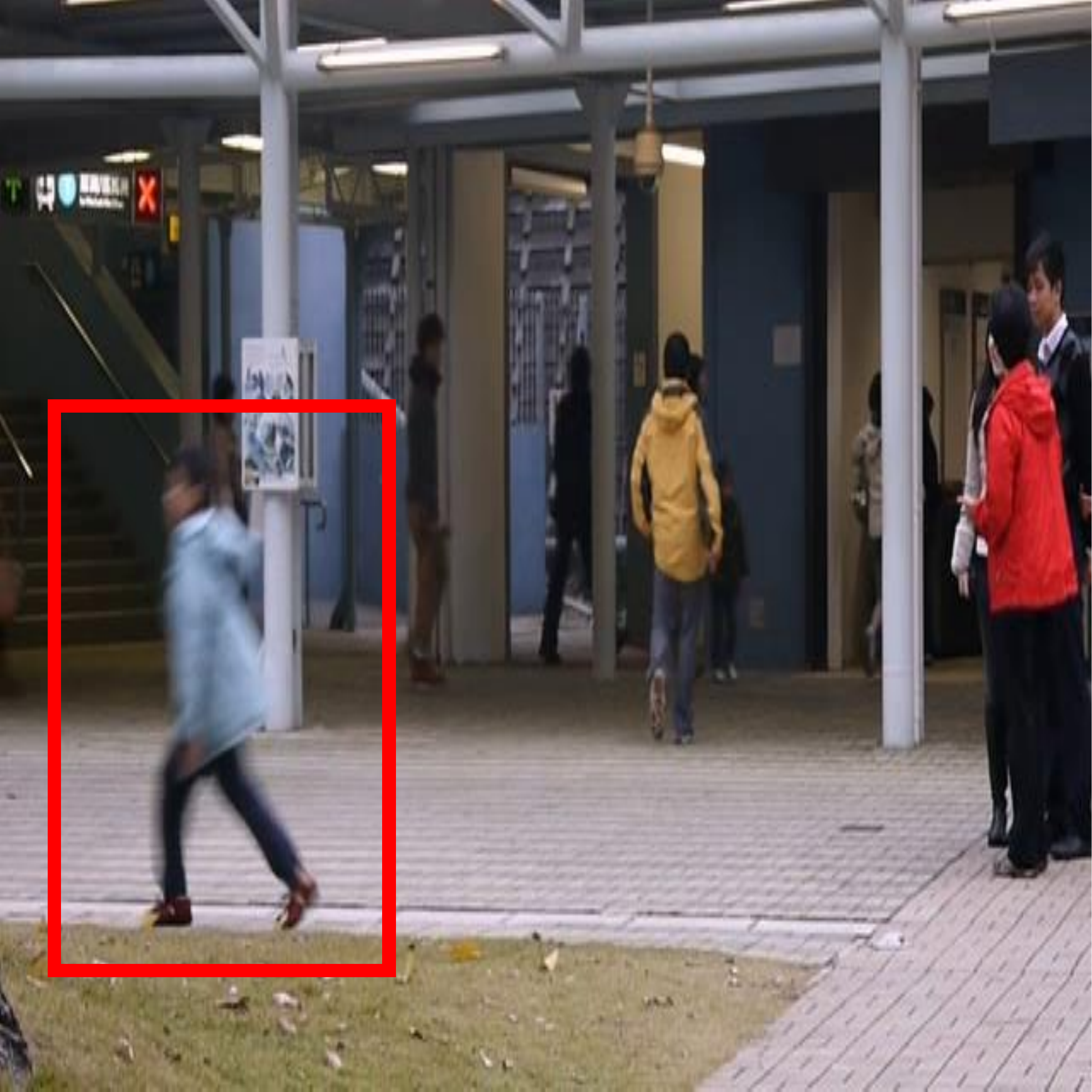} \label{f3_a}}
		\subfloat{\includegraphics[width=0.28\columnwidth]{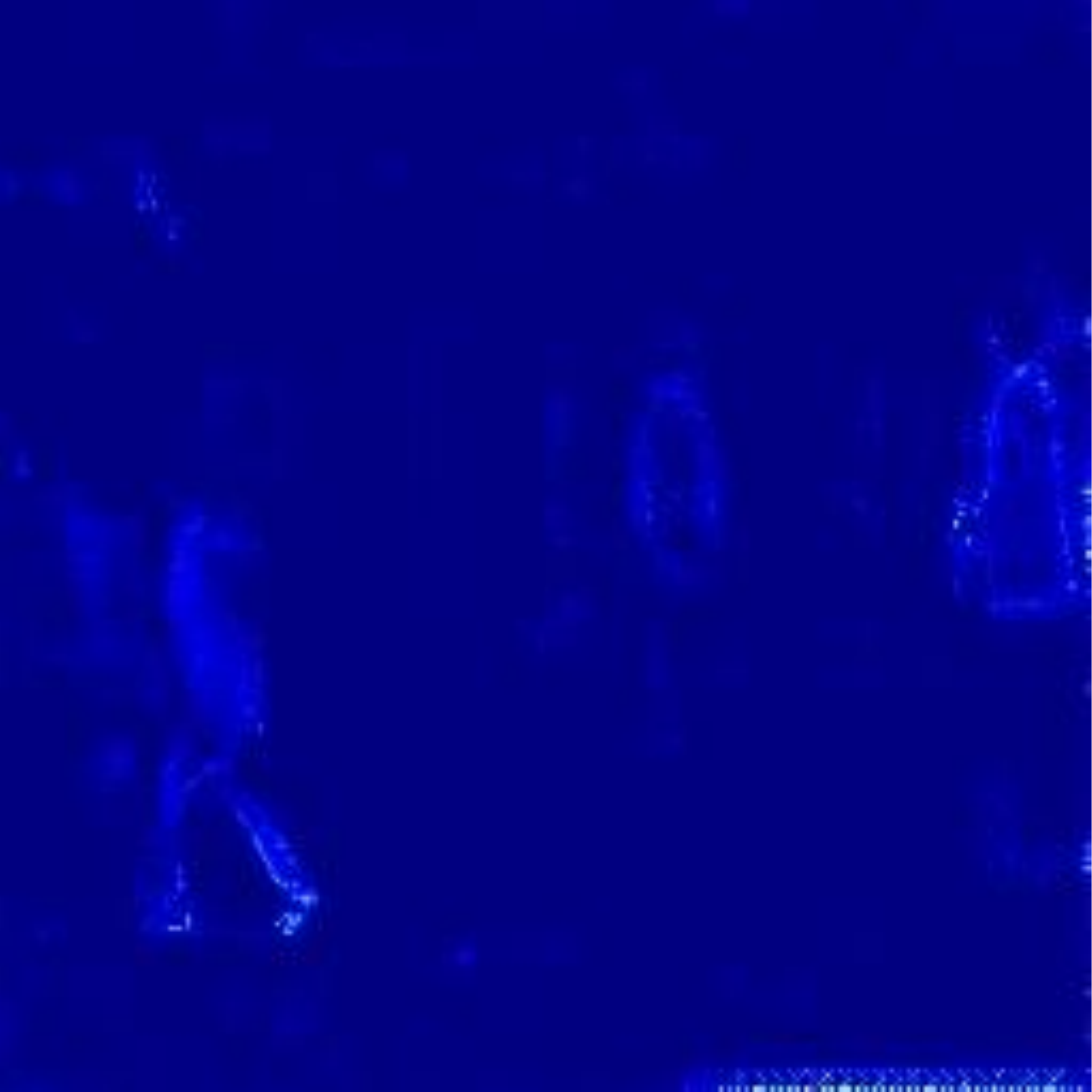}\label{f3_b}}
		\subfloat{\includegraphics[width=0.28\columnwidth]{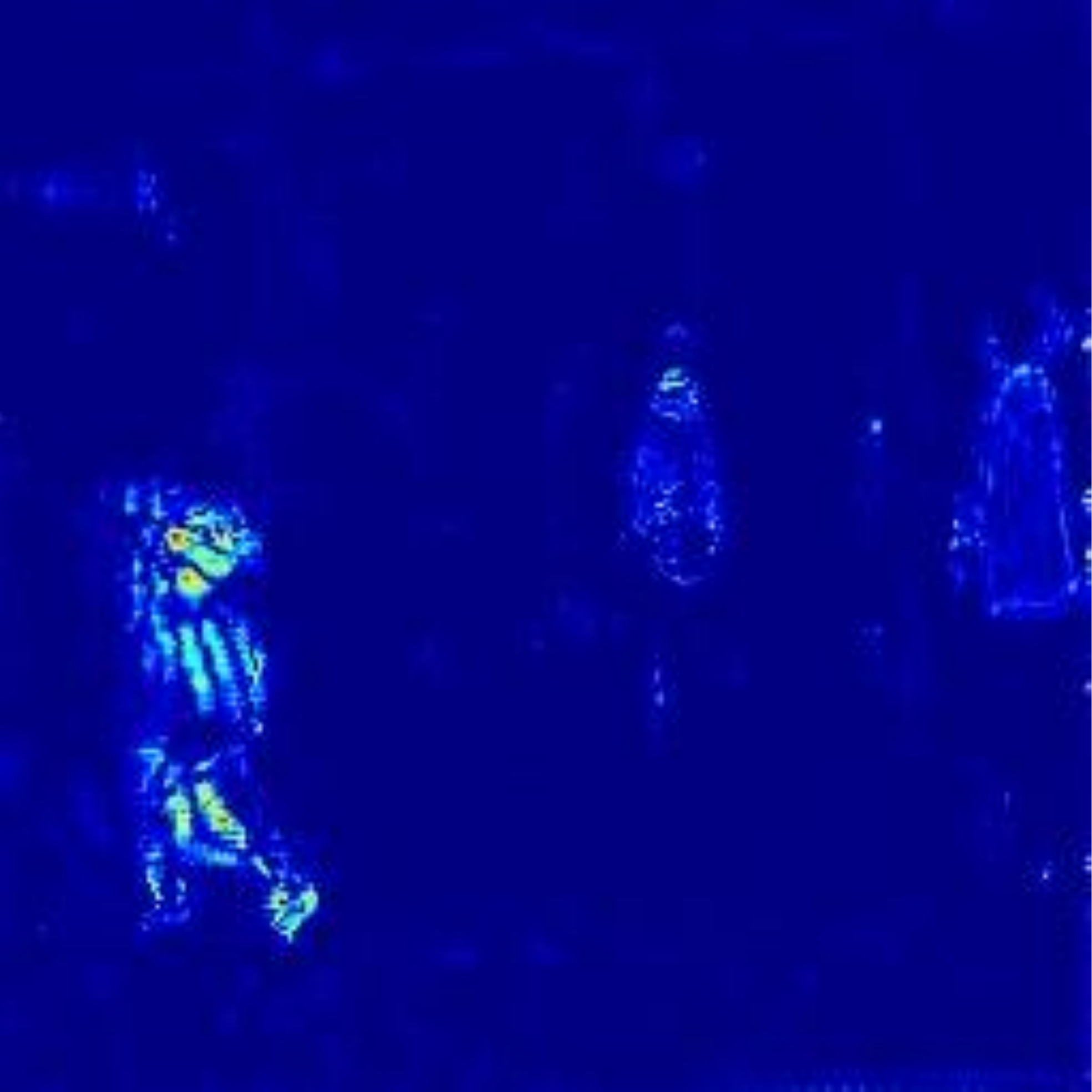}\label{f3_c}}
		\\
		\vspace{-0.3\baselineskip}
		\addtocounter{subfigure}{-3}
		\subfloat[Abnormal frame]{\includegraphics[width=0.28\columnwidth]{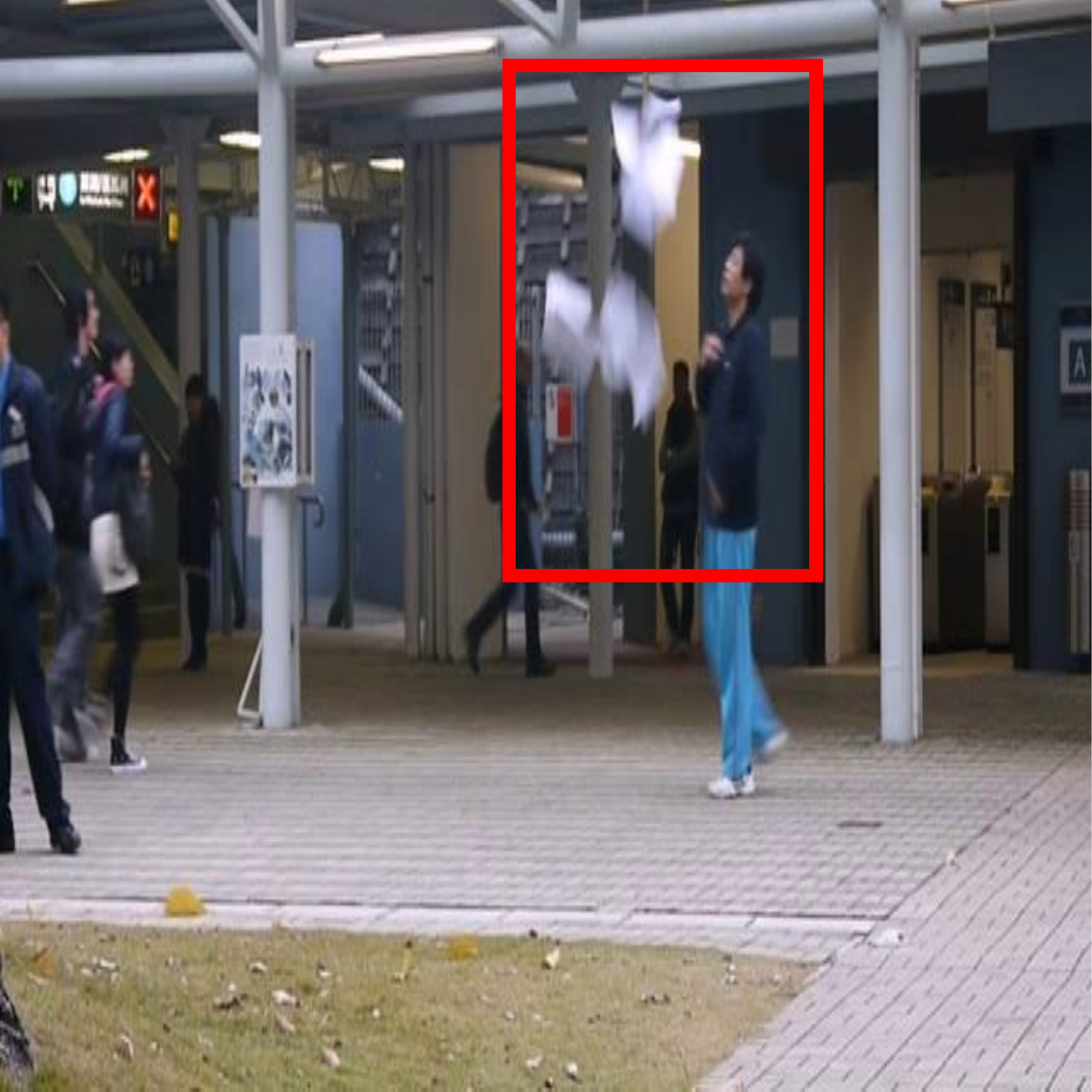}	\label{f3_a_2}}
		\subfloat[AE]{\includegraphics[width=0.28\columnwidth]{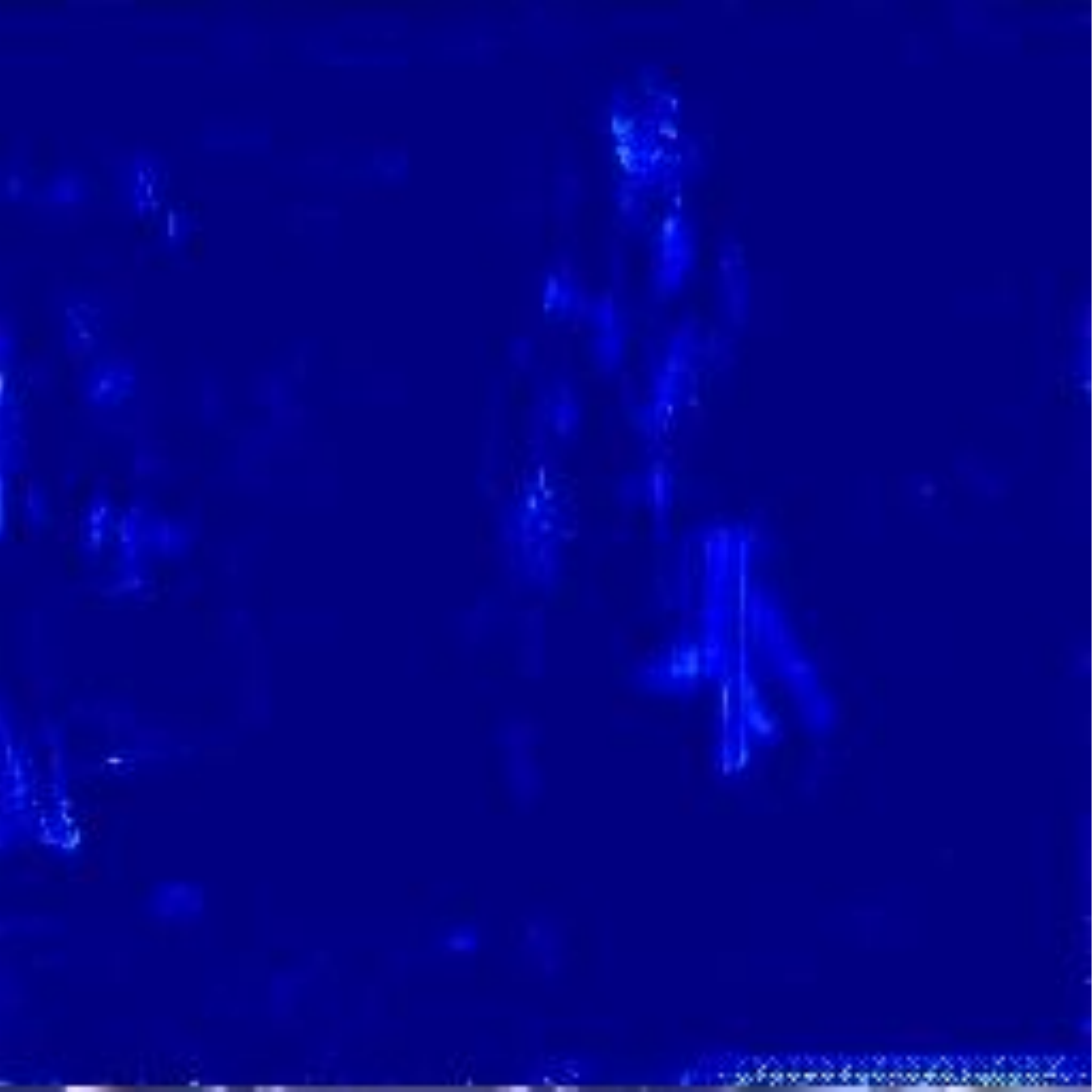}\label{f3_b_2}}
		\subfloat[ITAE]{\includegraphics[width=0.28\columnwidth]{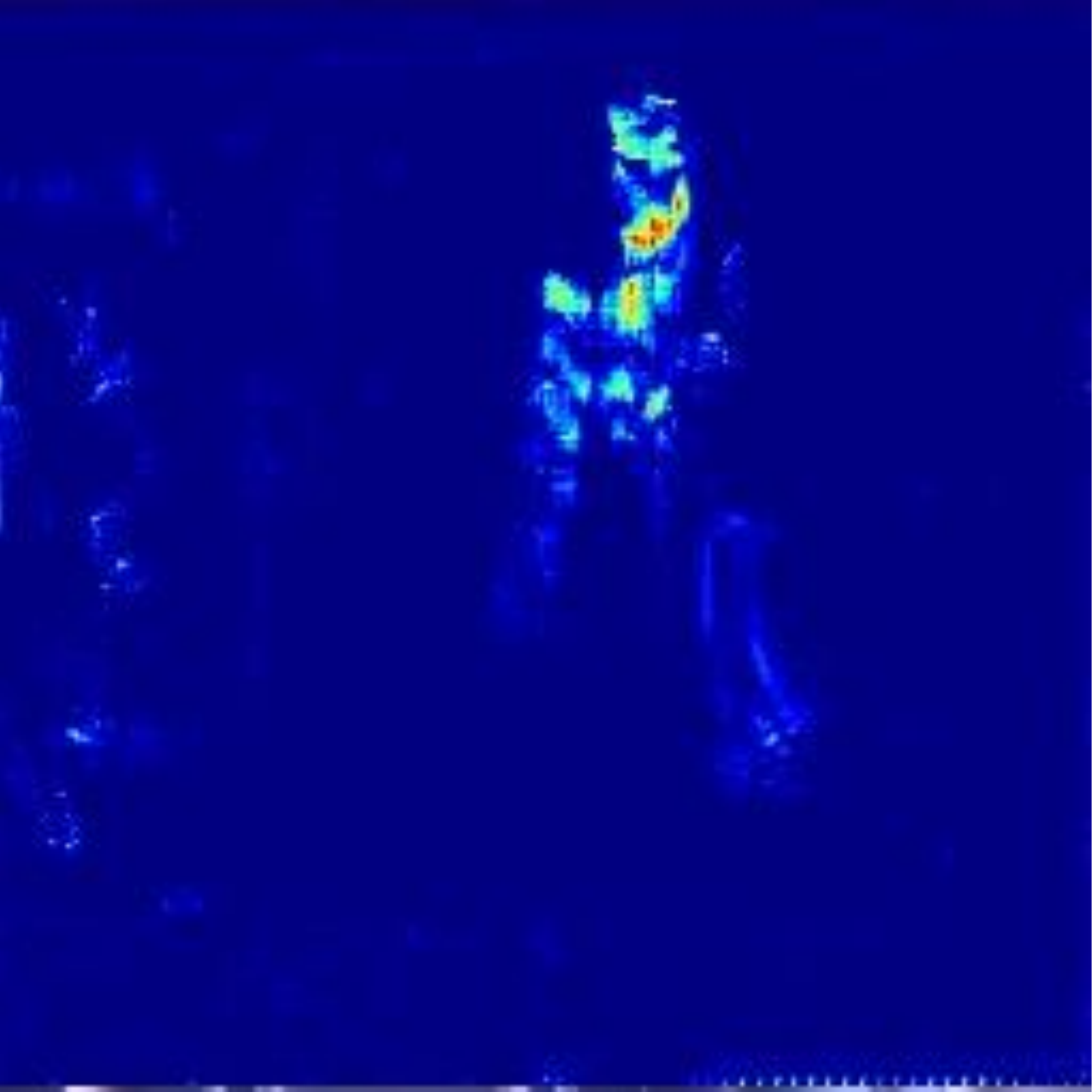}\label{f3_c_2}}
		\caption{Reconstruction error of AE and proposed ITAE in abnormal frames on CUHK Avenue database~\cite{lu2013abnormal}.}
		\label{f3}
	\end{figure}
	
	\subsection{Comparison}
	A comparison with other state-of-the-art methods on three benchmarks is presented in Table~\ref{t2}. The proposed approach achieves superior or competitive performance with the state-of-the-art methods on three datasets, and ITAE significantly improves performance over the well-designed AE (AE-Conv3D \cite{zhao2017spatio}, TSC \cite{luo2017revisit}, StackRNN \cite{luo2017revisit}, and DissociateAE~\cite{chang2022video}). In particular, ITAE and ITAE+NFs both show superior performance in UCSD Ped2 with 98.7\% and 99.2\% AUC, respectively, without any external models or datasets such as optical flow model (e.g., SelFlow, OpticalFlow, Flownet, Flownet2 pre-trained on FlyingThing3D, and ChairsSDHom datasets); object detector (e.g., FRCN, Yolov3 pre-trained on MS-COCO dataset); and feature extractor (e.g., Resnet-50 pre-trained on the ImagNet dataset).  
	In addition, compared to the latest methods using an object detector~\cite{georgescu2020background, ionescu2019object, georgescu2021anomaly}, some results show slightly higher performance than the proposed method, but the proposed method’s performance is still 0.5\%, 4.9\%, and 1.7\% higher on Ped2. Furthermore, these methods have a crucial issue: detecting anomalies is impossible for the object classes that are not pre-trained on external datasets as noted in BG-Agnostic~\cite{georgescu2020background}. Utilizing an object detector for knowledge distillation by considering unseen object classes (e.g., bike, car) during training as anomalies may be difficult to generalize in real-world scenarios. 
	
	Without an external model, the memory-based approach~\cite{gong2019memorizing, park2020learning} that stores and updates normal query features by memory module shows good performance, but this approach demonstrates low performance on the ST database with complex and multiple scenes, which indicates that storing fixed numbers of memory items may not be suitable for the general problem. In contrast, ITAE demonstrates better performance, i.e., 74.8\%, on the ST database, and the performance with NFs is 76.3\%, which shows the highest improvement in multi-scene anomaly detection among the results on the three benchmarks, which is very important from the perspective of generality. For all three databases, the results are superior to those of Auto-reg \cite{abati2019latent}, which performs auto-regressive generative modeling that requires a causal network and data ordering to perform sequential operations. It is noteworthy that the ITAE+NFs model  outperforms on three databases without using any external dataset and model, and NFs also show effectiveness on generalization through high performance improvement on multi-scene anomaly detection.
	
	\section{Discussions}
	\begin{figure}[!t]
		\centering
		\subfloat[UCSD Ped2~\cite{li2013anomaly}]{\includegraphics[width=0.95\columnwidth]{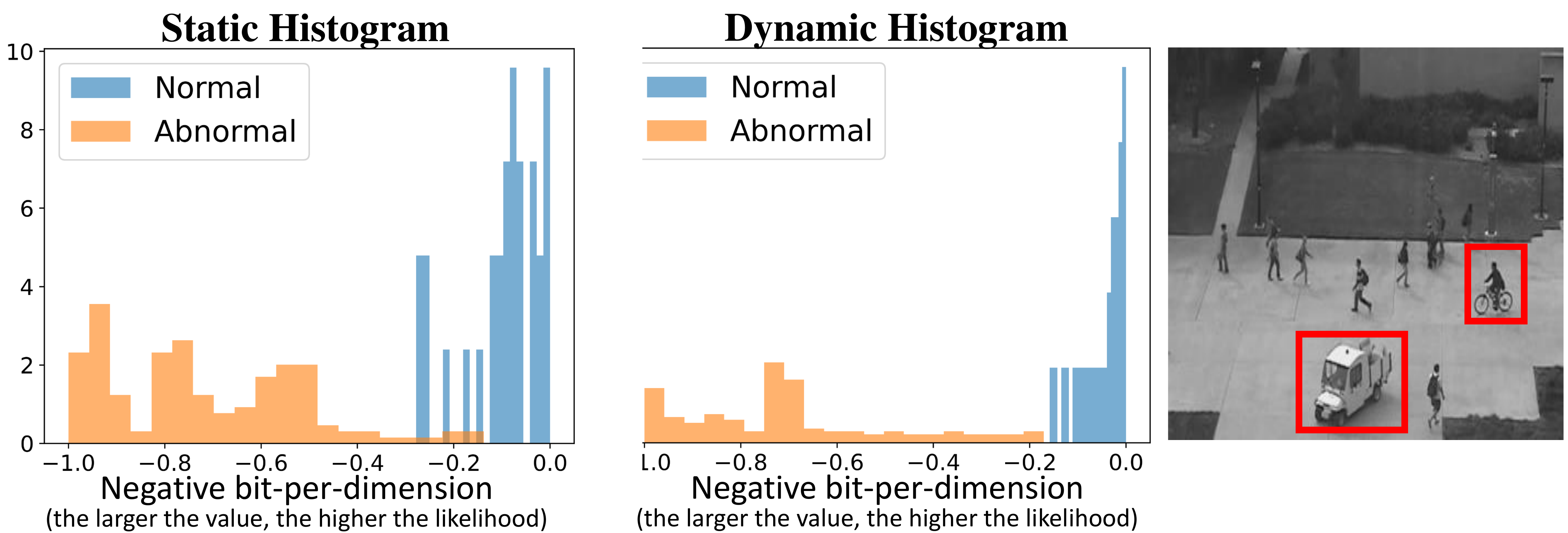} \label{f5_a}}
		\\
		\subfloat[Shanghai Tech~\cite{liu2018future}]{\includegraphics[width=0.95\columnwidth]{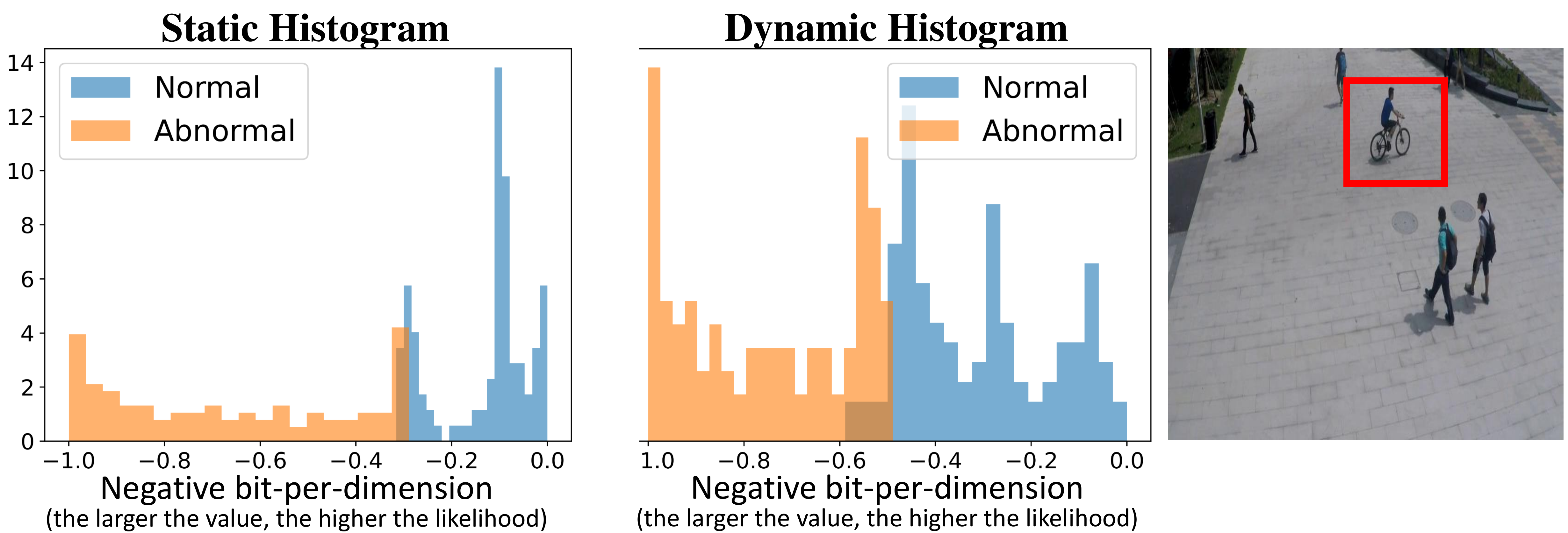}\label{f5_b}}
		\caption{Static and dynamic log-likelihood (expressed in normalized negative bits-per-dimension) histogram within a video clip.} 
		\label{f5}
	\end{figure}
	
	\begin{figure}[!t]
		\centering
		\subfloat[UCSD Ped2~\cite{li2013anomaly}]{\includegraphics[width=0.95\columnwidth]{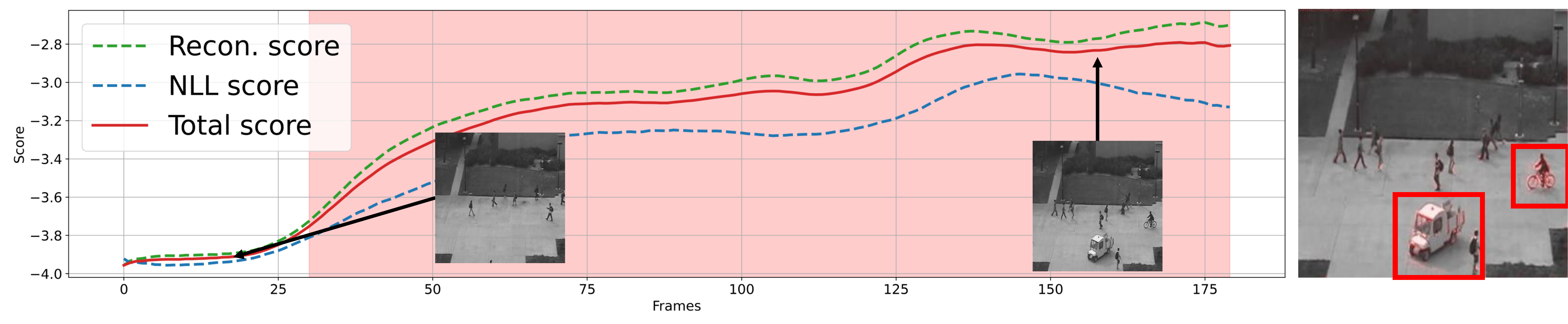} \label{f4_a}}
		\\
		\subfloat[CUHK~\cite{lu2013abnormal}]{\includegraphics[width=0.95\columnwidth]{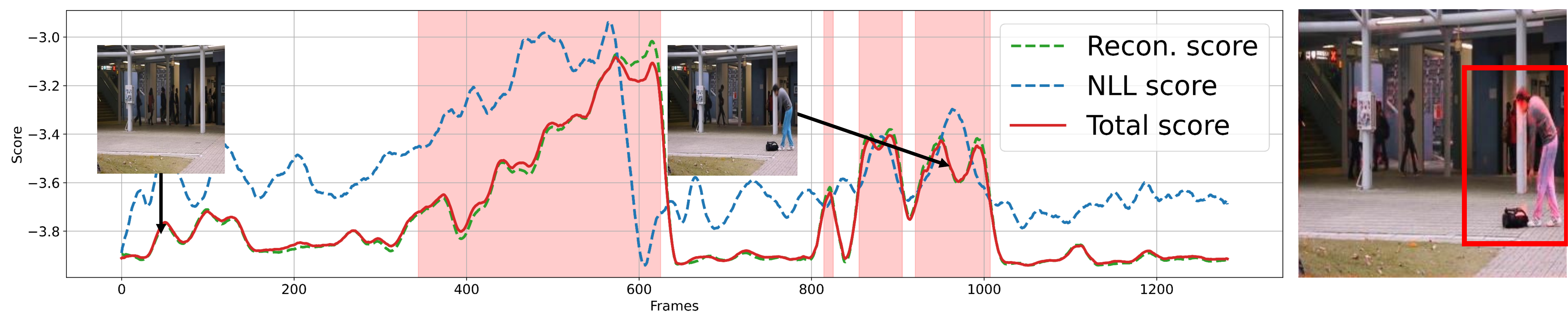}\label{f4_b}}
		\\
		\subfloat[Shanghai Tech~\cite{liu2018future}]{\includegraphics[width=0.95\columnwidth]{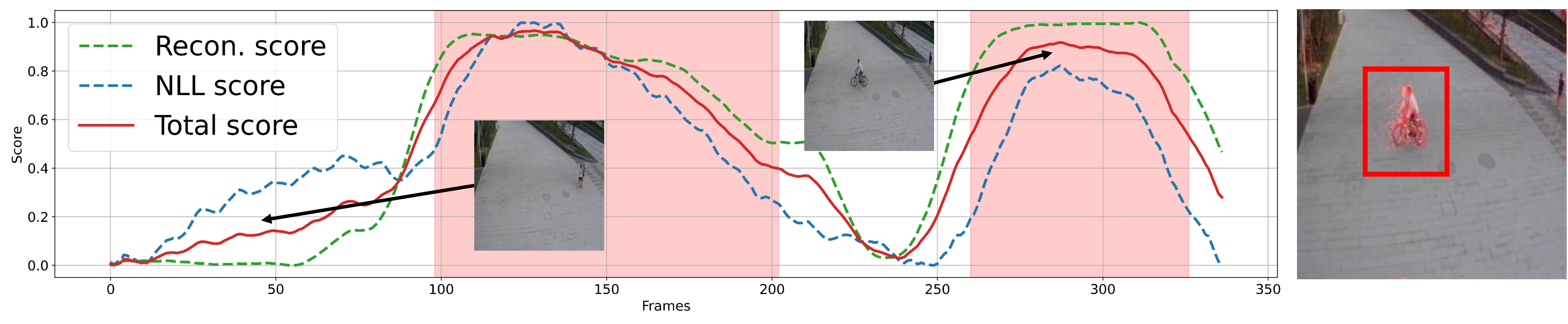}\label{f4_c}}
		\caption{Abnormal scores within a video clip and abnormal frames (errors are in red) in three benchmarks. The x-axis is the frame range and the y-axis is an anomalous score. The green, blue, and red lines indicate the ITAE score, NF model score, and total score, respectively. The red highlighted ranges are ground-truth abnormal frames.} 
		\label{f4} 
	\end{figure}
	
	\subsection{Qualitative Results}\label{visualization}
	Figure~\ref{f3} illustrates a comparison of AE (consisting of only a static encoder) and ITAE. As in a previous study~\cite{park2020learning}, the error maps are visualized by marking the pixel that is larger than the average error value within the frame. The first row of the figure is a jumping child whose appearance is normal, which leads to low reconstruction error with one-path AE. In contrast, the ITAE, which focuses on motion as well as appearance, generates a substantial error owing to the poor reconstruction for inputs that differ from the normal learned motion. For the second row, a person throwing a paper, the abnormality of the paper’s appearance, and flying motion produce a large error in ITAE. 
	
	Figure~\ref{f5} is a histogram of likelihood within a video clip from static and dynamic NF models. The first row (a) is a car and biker on the walkway, and the second row (b) is a person riding a bike on the sidewalk. As depicted in the histogram, the likelihoods of both NF models are low in abnormal frames, with abnormal appearance and motion on the sidewalk.
	
	With the two-path encoder and likelihoods, we compute the anomaly score of each frame in Fig.~\ref{f4}. The two scores complement each other and achieve satisfactory results, even when the pedestrian density is high or low and the foreground scale is small or large. (Please refer to the supplementary material for various qualitative results.)
	
	\begin{table}[!t]
		\centering
		\caption{Testing results on the cross-domain dataset. Each result is the performance on the target dataset of the model trained on the source dataset.}
		\resizebox{1.0\linewidth}{!}{
			{\Large
				\begin{tabular}{c|c|c|c|c}
					\hline
					\multicolumn{1}{c|}{\textbf{Source-Target}} & \multicolumn{2}{c|}{\textbf{Ours}} & \multirow{2}[1]{*}{\textbf{BG-Agnostic}~\cite{georgescu2020background}} & \multirow{2}[1]{*}{\textbf{Fewshot}~\cite{lu2020few}}\\
					\cline{2-3}\multicolumn{1}{c|}{\textbf{Dataset}}&\multicolumn{1}{c|}{\textbf{ITAE}}&\textbf{ITAE w NFs}&&\\
					\hline
					\hline
					Ped2 $\rightarrow$ Ped2 & 98.7&99.2 & 98.7 & 96.2\\
					CUHK $\rightarrow$ Ped2 & 97.0&94.3 \gray{(-4.9\%)}& 87.0 \gray{(-11.7\%)} & - \\
					ST $\rightarrow$ Ped2 & 97.8 & 96.8 \gray{(-2.4\%)} & 90.6 \gray{(-8.1\%)} & 82.0 \gray{(-14.2\%)}\\
					\hline
					CUHK $\rightarrow$ CUHK & 87.3& 88.0 & 92.3 & 85.8\\
					ST $\rightarrow$ CUHK &84.3 & 85.1 \gray{(-2.9\%)}& 83.6 \gray{(-8.7\%)}& 71.4 \gray{(-14.4\%)}\\
					\hline
					ST $\rightarrow$ ST & 74.8&76.3 & 82.7& -\\
					CUHK $\rightarrow$ ST &72.5& 73.0 \gray{(-3.3\%)}& 76.4 \gray{(-6.3\%)}& - \\
					\hline
					
			\end{tabular}}
		}
		\label{t6}
	\end{table}
	
	\subsection{Cross-domain Testing Results} \label{cross-domain}
	When considering real-world scenarios, generality is a crucial issue in surveillance anomaly detection. In order to detect undefined abnormal events, it is necessary to learn a general normal pattern through normal scenes of training data, and to prevent the occurrence of many false alarms by focusing on specific normality due to overfitting the data. As  the unsupervised anomaly detection approach trains with unlabeled data, we investigate the generalization ability by testing the model that trained the normal pattern on another dataset. As the Ped2 dataset is in grayscale , and the CUHK and ST datasets are composed of RGB scale frames, the color domain of the source and target datasets might differ. Instead of training a model with the same color domain as the target dataset, we use the model trained on the source dataset as it is. Therefore, when CUHK $\rightarrow$ Ped2 or ST $\rightarrow$ Ped2, given that the trained model on CUHK or ST takes 3-channel input, we duplicate the grayscale value to 3-channel for evaluation following Fewshot~\cite{lu2020few}.
	
	Table~\ref{t6} presents the result of testing the model, which was trained on the source dataset, on the target dataset. ITAE and ITAE+NFs both show a slight decrease in performance, but this decrease is much smaller than the latest methods, i.e., BG-Augnostic~\cite{georgescu2020background} and Fewshot~\cite{lu2020few}. When the target dataset is Ped2 and the source dataset is CUHK or ST, the performance degradation of the proposed method is 4.9\% and 2.4\%, respectively, that of BG-Augnostic is 11.7\% and 8.1\%, respectively, and that of Fewshot is 14.2\% for CUHK. ITAE and NFs also show the smallest decrease in other target datasets, and the highest performance in Ped2 and CUHK datasets. The degradation is the most in CUHK $\rightarrow$ Ped2, because Ped2 is grayscale unlike CUHK and the characteristics of the dataset such as camera angle and object size differ the most. For this reason, the distribution modeling of NFs is difficult owing to the large difference between the target and source datasets and the performance of ITAE+NFs is lower than that of ITAE when Ped2 is the target dataset. ITAE+NFs, which focuses on static and dynamic features and performs normality learning, shows high performance in cross-domain testing and effectiveness in generalization ability.
	
	\begin{figure}[!t]
		\centering
		\includegraphics[width=0.7\linewidth]{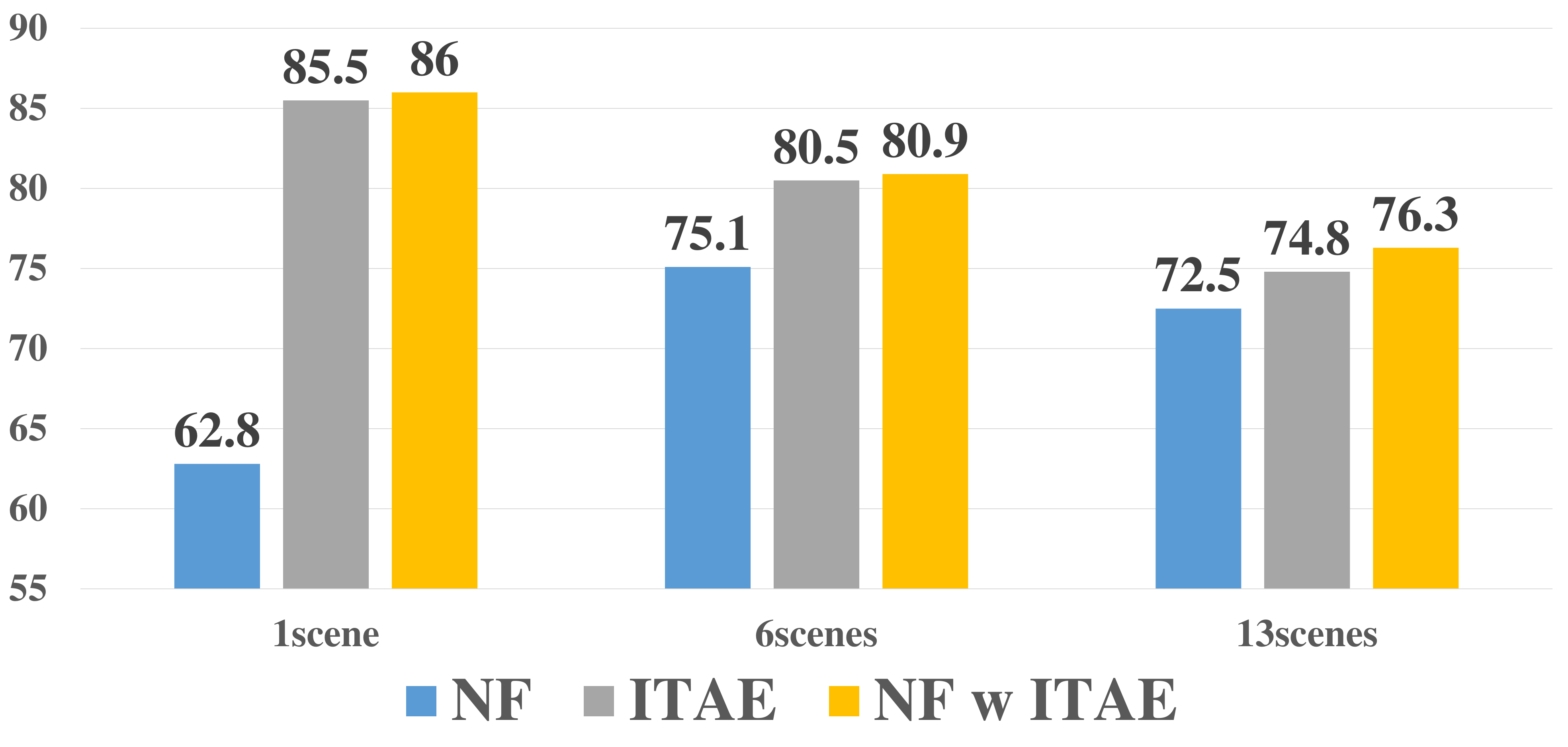}
		\caption{Performance of ITAE and NF models for 1, 6, and 13 scenes in ST database.}
		\label{f6}
	\end{figure}
	
	\begin{table}[!t]
		\centering
		\caption{Ablation studies of ITAE and NF models on three benchmark databases.}
		\resizebox{1.0\linewidth}{!}{
			\renewcommand{\arraystretch}{1}
			{\Huge
				\begin{tabular}{cccc|cc|cc|cc}
					\hline
					\multicolumn{2}{c}{\textbf{ITAE}} & \multicolumn{2}{|c|}{\textbf{NFs}}&
					\multicolumn{2}{c|}{\textbf{(a) UCF-Crimes}~\cite{sultani2018real}} & \multicolumn{2}{c|}{\textbf{(b) LV}~\cite{leyva2017lv}}&
					\multicolumn{2}{c}{\textbf{(c) UBI-Fights}~\cite{degardin2020human}}\\
					\cline{1-4}\multicolumn{1}{c}{\textbf{Static}} & \multicolumn{1}{c|}{\textbf{Dynamic}} & \multicolumn{1}{c}{\textbf{Static}} & \multicolumn{1}{c|}{\textbf{Dynamic}} & \textbf{AUC} & \textbf{EER }& \textbf{AUC} & \textbf{EER}& \textbf{AUC} & \textbf{EER} \\
					\hline
					\hline
					\checkmark& & & & 66.0 &36.7 &70.6&36.9&56.2 &45.4 \\
					\checkmark&\checkmark& & & 66.3 &36.9 &74.6&30.2&57.7 &44.2 \\
					& &\checkmark&\checkmark& 67.2&38.3 &77.1 & 29.2& 53.3&48.0\\
					\checkmark&\checkmark&\checkmark& & 69.7&36.7 & 77.2&27.6& 57.9&44.3 \\
					\checkmark&\checkmark& &\checkmark& 70.7&34.9 & 77.4&28.3& 57.8&44.1 \\
					\checkmark&\checkmark&\checkmark&\checkmark& 70.9&35.2 & 77.5&27.8& 57.8&44.1 \\
					\hline
				\end{tabular}
				
			}
		}
		\label{t_real}
	\end{table}	
	\begin{table*}[!t]
		\centering
		\caption{Comparison with unsupervised anomaly detection methods on three benchmarks; comparison result of our model is the performance of ITAE and NFs. DEC indicates decidability index, which measures how well intra-class (genuine) and inter-class (impostor) distribution scores are distant from each other.}
		\begin{subtable}[t]{0.3\linewidth}
			\centering
			\caption{UCF-Crimes~\cite{sultani2018real}}
			\resizebox{0.95\linewidth}{!}{
				\begin{tabular}{c|c|c}
					\hline
					\textbf{Methods} & \multicolumn{1}{c|}{\textbf{Features}} & \multicolumn{1}{l}{\textbf{AUC}} \\
					\hline
					\hline
					SVM Baseline &    -   & 50.0 \\
					ConvAE~\cite{hasan2016learning} &   -    & 50.6 \\
					S-SVDD~\cite{sohrab2018subspace} &   -    & 58.5 \\
					SCL~\cite{lu2013abnormal}    & \multicolumn{1}{c|}{C3D RGB} & 65.5 \\
					BODS~\cite{wang2019gods}  & \multicolumn{1}{c|}{I3D RGB} & 68.3 \\
					GODS~\cite{wang2019gods}  & \multicolumn{1}{c|}{I3D RGB} & 70.5 \\
					\hline
					ITAE (ours) &   -    & 66.3 \\
					ITAE+NF (ours) &    -   & \textbf{70.9} \\
					\hline
				\end{tabular}
				\label{t_real2_1}
			}
		\end{subtable}%
		\begin{subtable}[t]{0.25\linewidth}
			\centering
			\caption{Live-Video~\cite{leyva2017lv}}
			\resizebox{0.95\linewidth}{!}{
				\begin{tabular}{c|cc}
					\hline
					\textbf{Method} & \multicolumn{1}{c}{\textbf{AUC}} & \multicolumn{1}{c}{\textbf{EER}} \\
					\hline
					\hline
					AME~\cite{gunale2018deep} &   39.8   & 57.2\\
					SCL~\cite{lu2013abnormal} &   49.6  & 51.0 \\
					MVs~\cite{biswas2013real}   & 56.6 & 44.9 \\
					CS~\cite{leyva2017video}  &61.8 & 41.0 \\
					DeepOC~\cite{wu2019deep}  & 70.6 & 35.1 \\
					\hline
					ITAE (ours) &   74.6  & 30.2 \\
					ITAE+NF (ours) &  \textbf{77.5}  & \textbf{27.8} \\
					\hline
				\end{tabular}
				\label{t_real2_2}
			}
		\end{subtable}%
		\begin{subtable}[t]{0.33\linewidth}
			\centering
			\caption{UBI-Fights~\cite{degardin2020human}}
			\resizebox{0.95\linewidth}{!}{
				\begin{tabular}{c|ccc}
					\hline
					\textbf{Method} & \multicolumn{1}{c}{\textbf{AUC}} & \multicolumn{1}{c}{\textbf{EER}}& \textbf{DEC} \\
					\hline
					\hline
					spatiotemporal~\cite{chong2017abnormal} &  52.8&44.6&\textbf{0.194}\\
					LTR~\cite{hasan2016learning} &   53.3&48.4&0.147\\
					abnormalGAN~\cite{ravanbakhsh2017abnormal}   &54.0&47.5&0.164\\
					$S^2$ VAE~\cite{wang2018generative}  &54.1&48.0&0.059 \\
					Binary SVM  & 55.6 & 44.3&0.128 \\
					\hline
					ITAE (ours) &   57.7  & 44.2&0.098 \\
					ITAE+NF (ours) & \textbf{57.8}&\textbf{44.1}&0.102 \\
					\hline
				\end{tabular}
				\label{t_real2_3}
			}
		\end{subtable}%
		\label{t_real2}
	\end{table*}%
	
	\subsection{Analysis According to the Number of Scenes} \label{multiscene}
	We compose subsets with 1, 6, and 13 scenes in the ST dataset and compare the performance as the scenes of the data become diverse and extensive. The sizes of the test sets are proportional, and the train-to-test set ratio in each subset is similar. In Fig.~\ref{f6}, through various scenes and a large number  of videos, the NF model learns the general normal pattern and shows higher performance on a multiple scene subset than on a single scene subset. In contrast, ITAE illustrates higher performance as there are fewer scenes because of overfitting, but exhibits limitations as scenes become more diverse. NFs complement the drawback of AE and produce an excellent performance improvement (1.5\%) in the most diverse 13-scene subset. From Fig.~\ref{f6} and Table~\ref{t5}, it can be observed that the distribution learning of the NF models on ITAE features is effective in coping with the complex normality distribution.
	
	\subsection{Experimental Result and Analysis on Real-world Scenario Benchmarks}
	We also consider more realistic and dynamic databases: UCF-Crimes, LV, and UBI-Fights. As our method is an unsupervised-learning-based approach, the framework is trained using only normal videos from the training set of each database. In Table~\ref{t_real} (a) and (b), the performance NFs is higher than that of ITAE in UCF-Crimes as well as LV datasets. In Section~\ref{multiscene}, the experimental results show that as the number of scenes increases, the performance of the autoencoder deteriorates, while the performance of NF model remains similar by learning the distribution of general normal patterns. In the case of UBI-Fights, in Table\ref{t_real}~(c), ITAE and NFs achieve 57.7\% and 53.3\% AUC score, respectively. ITAE shows better performance than NFs, which seems to indicate that detecting anomalies with reconstruction error is more effective than distinguishing abnormality by distribution of latent features when the abnormal event is fighting; this is because a fighting event involves a large change in appearance and motion at the frame level, which results in a large reconstruction error that leads to anomaly detection. In Table~\ref{t_real2}, the performance of ITAE with NFs on UCF-Crimes, LV, and UBI-Fights dataset is 70.9\%, 77.5\%, and 57.8\% respectively, and achieves the best results compared with unsupervised anomaly detection methods.
	
\begin{figure}[!t]
		\centering
		\subfloat{\includegraphics[width=0.22\columnwidth]{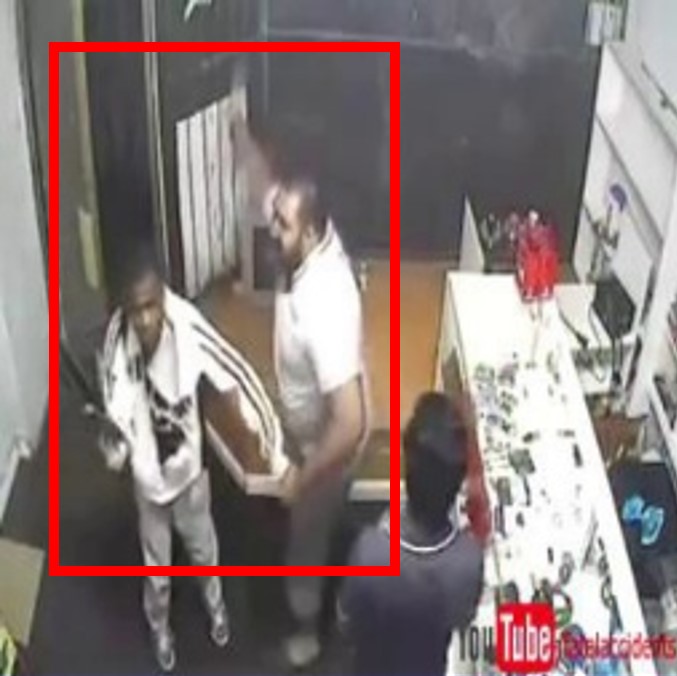}}
		\subfloat{\includegraphics[width=0.22\columnwidth]{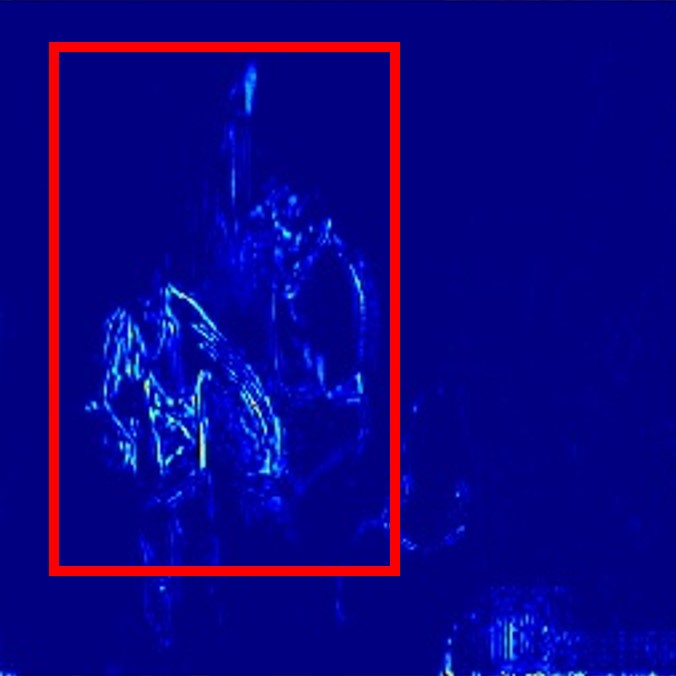}
		}
		\subfloat{\includegraphics[width=0.52\columnwidth]{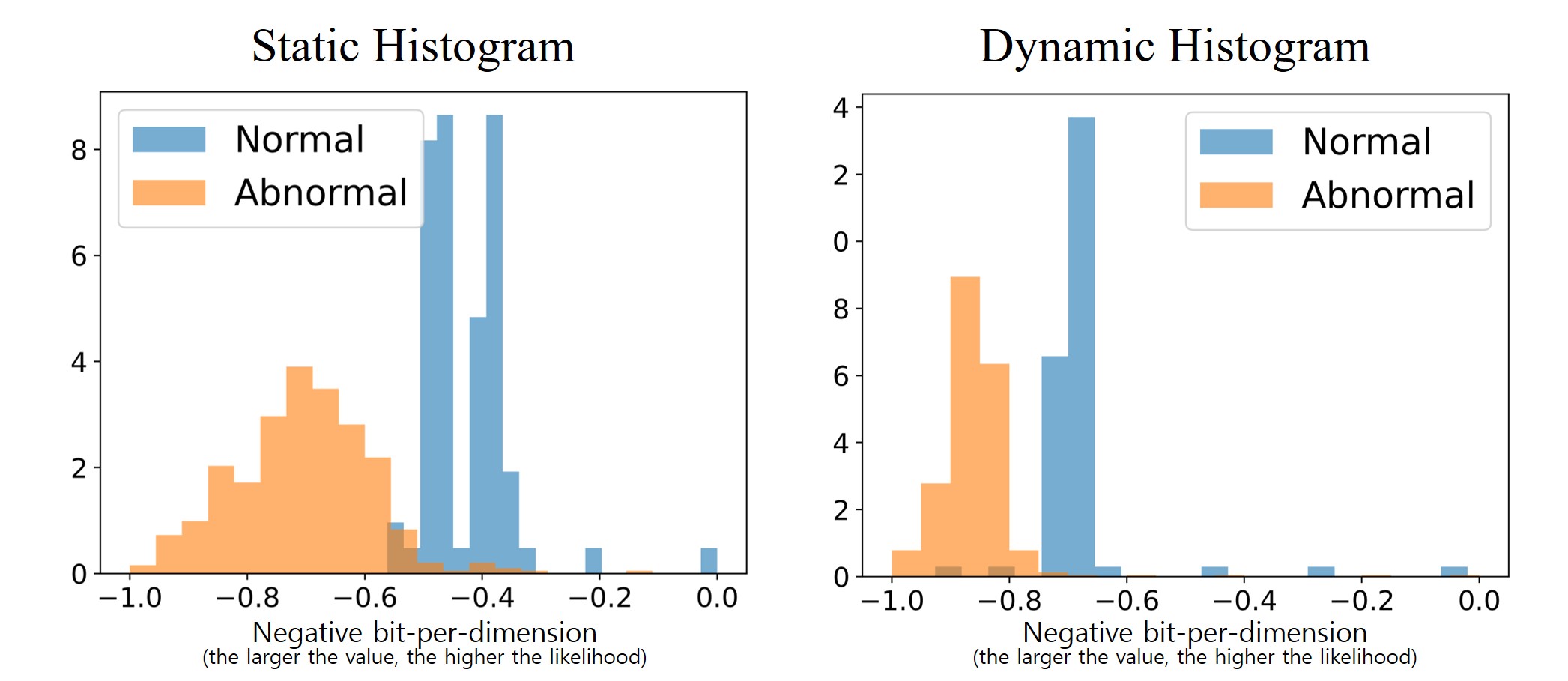}
		}\\
		\addtocounter{subfigure}{-3}
		\subfloat[Anomalies]{\includegraphics[width=0.22\columnwidth]{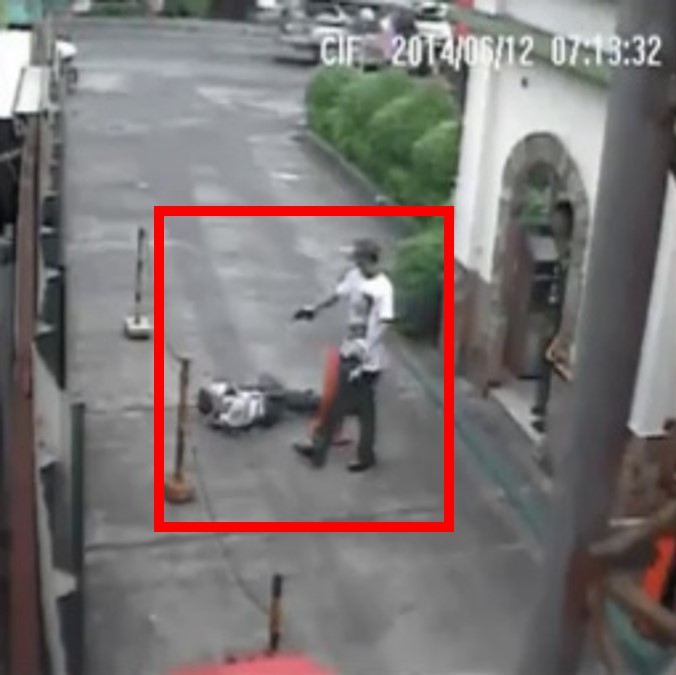}}
		\subfloat[Error map]{\includegraphics[width=0.22\columnwidth]{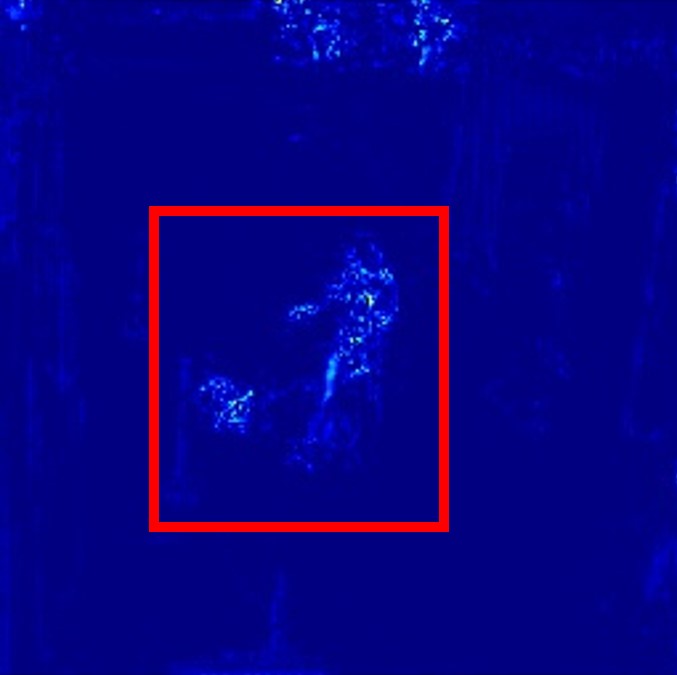}
		}
		\subfloat[Likelihoods from NFs]{\includegraphics[width=0.52\columnwidth]{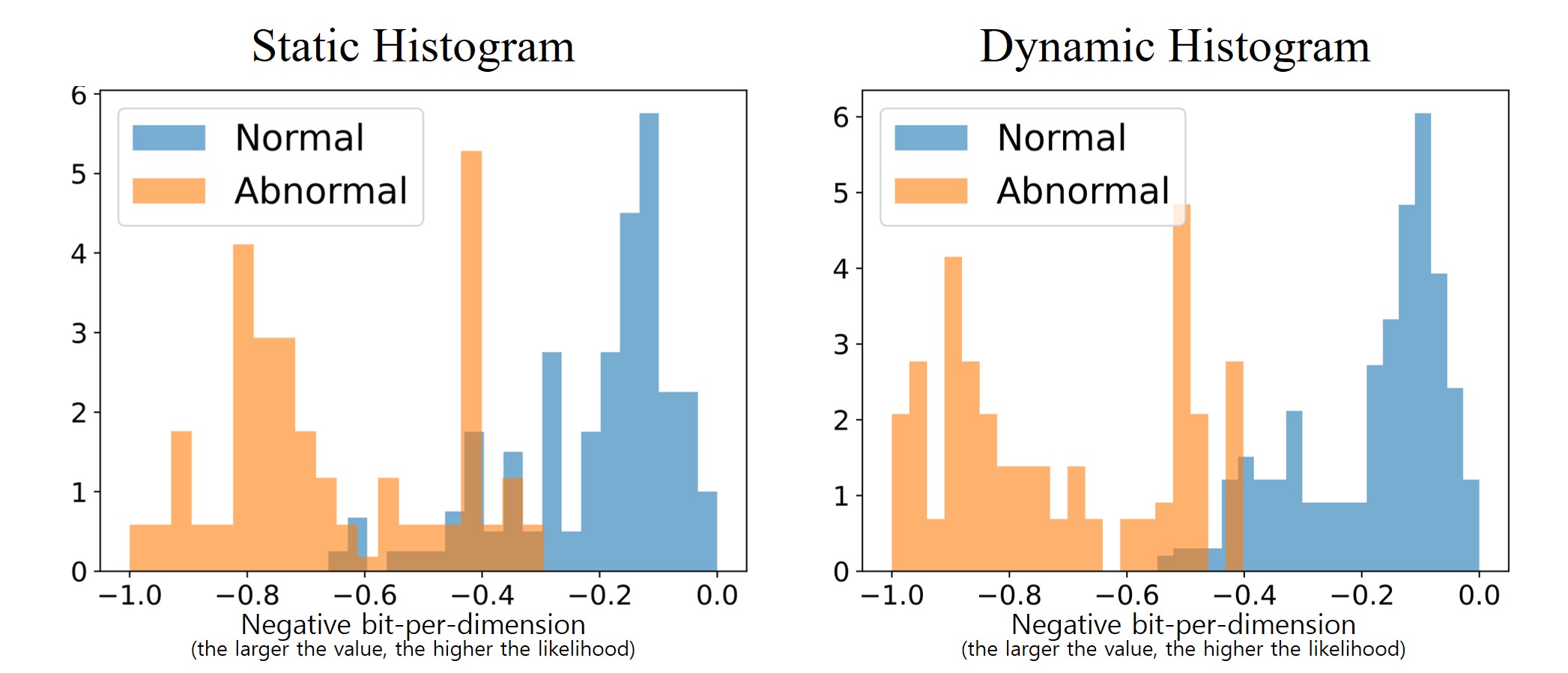}
		}
		\caption{Qualitative results of ITAE and NFs. Each column is the visualization of (a) abnormal frame, (b)error map from ITAE, and (c) static and dynamic log-likelihood (expressed in normalized negative bits-per-dimension) histogram within a test video clip.}
		\label{f_real}
	\end{figure}

	In Fig.~\ref{f_real}, the first row is an example of the UCF-Crimes dataset `Robbery048' clip, which is a scene of one person assaulting another; the second row is the LV dataset ‘murder’ clip, which is a scene showing a person being shot. Qualitative results show that anomalies are well distinguished even in clips of complex and diverse anomalies in real-world scenarios.
	
	\begin{figure}[!t]
		\centering
		\subfloat[False positive sample of AE]{\includegraphics[width=0.48\columnwidth]{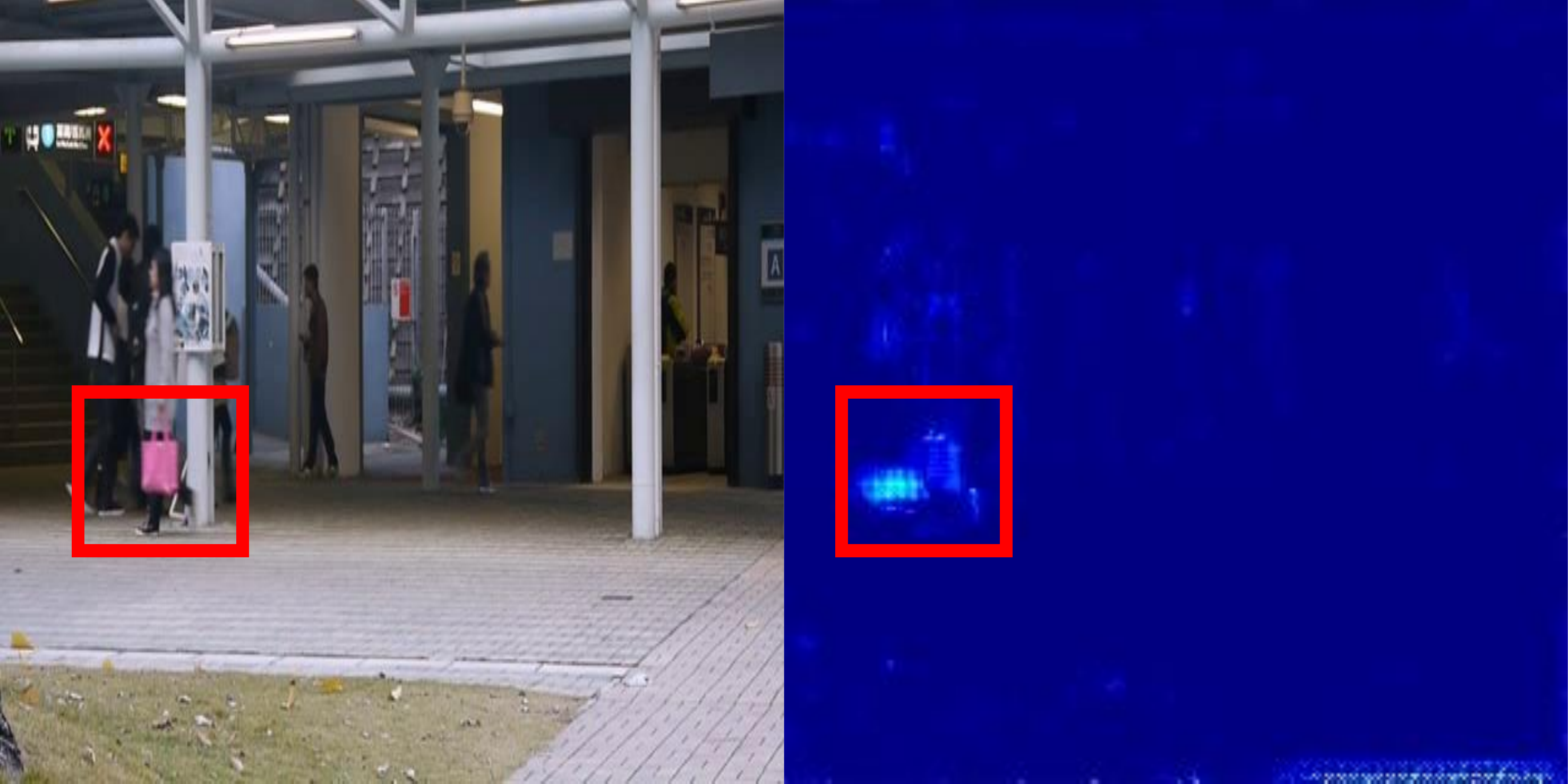} \label{f8_a}}
		\hfil
		\subfloat[False positive sample ITAE]{\includegraphics[width=0.48\columnwidth]{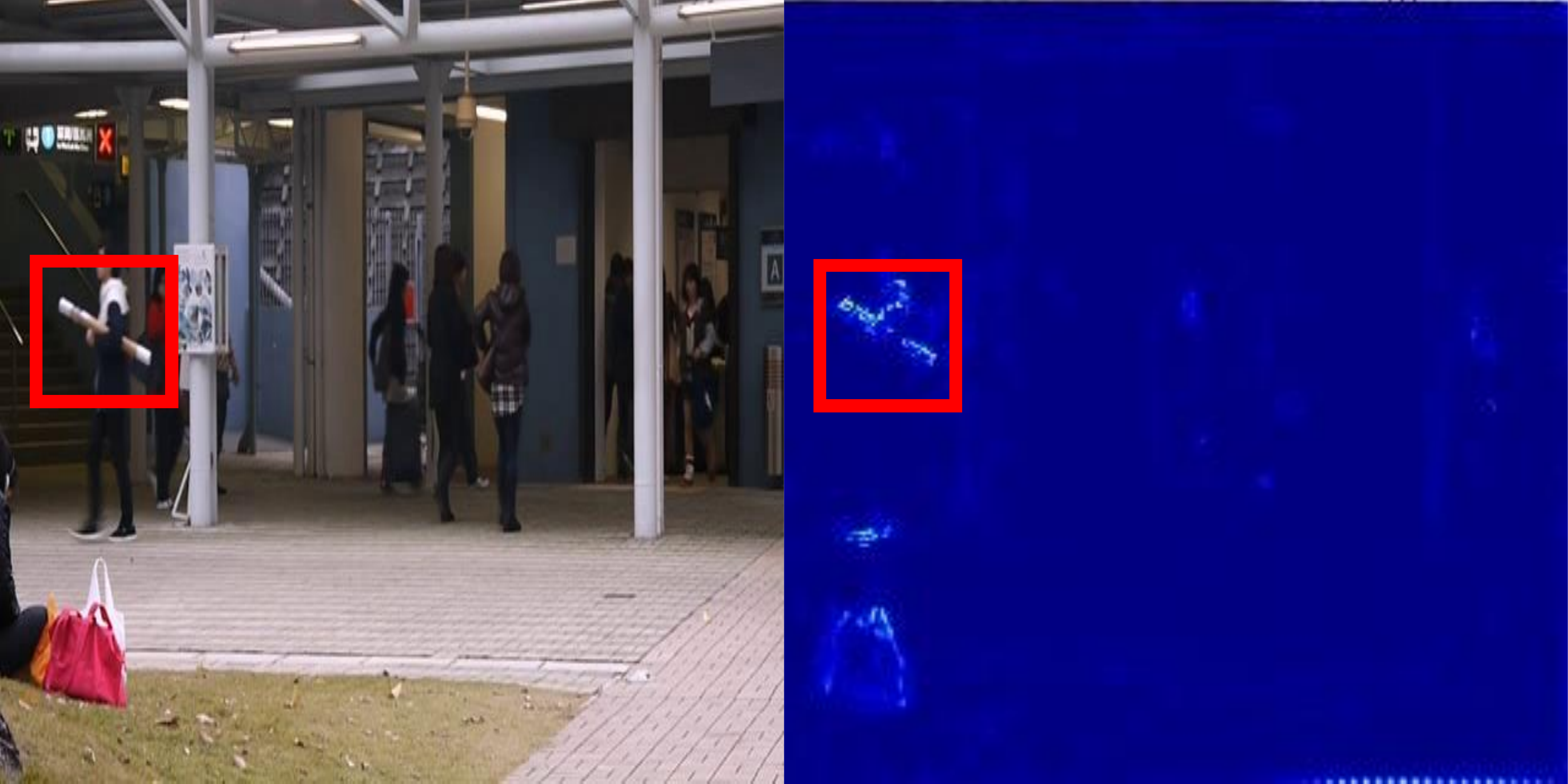}\label{f8_b}} 
		\\
		\subfloat[False negative sample ITAE w NF]{\includegraphics[width=1\columnwidth]{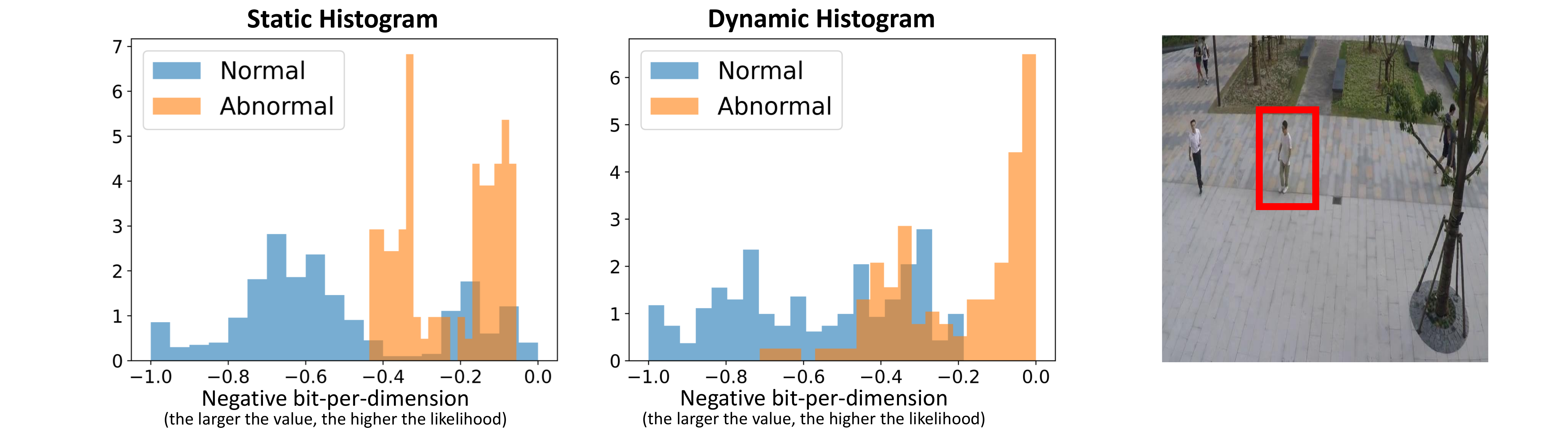}	\label{f8_c}}
		\caption{Samples of false positive and false negative scenes. (a) is a false positive sample only in AE, and (b) is a false positive sample in ITAE but is true negative in ITAE with NFs. (c) is a false negative sample in all cases.}
		\label{f8}
	\end{figure}
	
	\subsection{False Positive and Negative Samples}
	Figure~\ref{f8} shows false positive (FP) and false negative (FN) samples. In (a), unlike ITAE which considers appearance and motion factor concurrently, the one-path AE shows a high reconstruction error in the pedestrian carrying the red bag. (b) is an FP sample of ITAE, a scene in which a pedestrian carries a long stick, showing a high error on the strange stick that causes a high anomaly score, but the false alarm is prevented by adding the score of the NF model. (c) is an FN sample of the proposed method, a scene where a pedestrian is loitering. In this scene, the appearance of a pedestrian and the motion of walking are both similar to the normal pattern, so the ITAE and NFs are unable to distinguish between normal and abnormal scenes. 
	
	\subsection{Running Time and Model Complexity}
	We measure the computational time of the proposed method on the UCSD Ped2 dataset (resolution: $240\times360$) following~\cite{georgescu2021anomaly}. ITAE infers the anomaly score of a frame in 9.1 milliseconds (ms), while static and dynamic NF models take 22.2 ms and 20.7 ms, respectively. The total framework takes 52 ms per frame, while calculating the anomaly score takes 0.5 ms. With all components together, the proposed method runs at 19 FPS on an NVIDIA GeForce RTX 3090, 24GB memory. The parameter number of ITAE, static NFs, and dynamic NFs is 7.74, 10.16, and 11.05M, respectively; compared to one-path AE of 5.76M, ITAE with motion path only increases 1.98M parameters as the dynamic encoder focuses on motion with a high frame-rate and low channel size (1/8 of static).
	
	\section{Conclusion}
	In this paper, we proposed an ITAE and the distribution modeling of normal features based on an NF model in an unsupervised manner for video anomaly detection. We designed the ITAE to implicitly capture representative static and dynamic information of normal scenes without using a pre-trained network. For the complex normality, using the latent features of ITAE, we modeled the distribution of appearance and motion normality using an NF model through the tractable likelihood. In an experiment on standard benchmarks, ITAE demonstrated high effectiveness in scenes where motion is abnormal by learning the dynamic information of normal scenes. Furthermore, the normality modeling of the ITAE feature achieved superior results, especially when the database is extensive and composed of diverse scenes. The proposed method can be expected to model a general distribution and solve practical problems through a vast number of real-world videos with unsupervised learning. 
	
	As CCTVs exist in most places and keep records of our daily lives, their use may involve ethical concerns related to privacy invasion. However, from the perspective of the development of computer vision and pattern recognition applications, as surveillance anomaly detection can help to quickly detect accidents and crimes or automatically prevent them in advance, the reduction of human time, labor, and positive social impact is noteworthy. Our work, which is an unsupervised approach, makes it possible to learn general normal patterns using various scenarios of real-world surveillance video, which is promising and expected to accelerate detecting anomalies in contemporary society.
	
	\vspace{5mm}
	\noindent
	\textbf{Acknowledgement} 
	\newline
	This research was supported by Multi-Ministry Collaborative R\&D Program (R\&D program for complex cognitive technology) through the National Research Foundation of Korea (NRF) funded by MSIT, MOTIE, KNPA (NRF-2018M3E3A1057289)
	
	{\small
		\bibliographystyle{ieee_fullname}
		\bibliography{egbib}
	}

\end{document}